\documentclass{article}
\usepackage[final]{neurips_2024}
 \usepackage{float}
\usepackage{amsmath}
\usepackage[linesnumbered,ruled,vlined]{algorithm2e}

\usepackage{amsmath,amssymb,amsthm,bbm}
\usepackage[normalem]{ulem}
\usepackage[dvipsnames]{xcolor}
\usepackage{tikz-qtree}
\usepackage{endnotes}
\usepackage{graphicx}
\usepackage{endnotes}
\usepackage{fancyhdr}
\usepackage{datetime}
\usepackage{forest}
\usepackage{mdframed}
\usepackage{svg}
\usepackage{wrapfig}

\usepackage[utf8]{inputenc} 
\usepackage[T1]{fontenc}    
\usepackage{hyperref}       
\usepackage{url}            
\usepackage{booktabs}       
\usepackage{amsfonts}       
\usepackage{nicefrac}       
\usepackage{microtype}      

\usepackage{pifont}

\newcommand{\singbr}[1]{\left[#1\right]}
\newcommand{\myset}[1]{\left\{#1\right\}}
\newcommand{\paren}[1]{\left(#1\right)}
\newcommand{\expect}[1]{\E\singbr{#1}}

\newcommand{\norm}[1]{\left|\left|#1\right|\right|}

\newcommand{\tr}[1]{\mathrm{tr}\paren{#1}}

\DeclareFontFamily{U}{mathx}{\hyphenchar\font45}
\DeclareFontShape{U}{mathx}{m}{n}{<-> mathx10}{}
\DeclareSymbolFont{mathx}{U}{mathx}{m}{n}
\DeclareMathAccent{\widebar}{0}{mathx}{"73}
\newcommand*\wb[1]{\ensuremath{\widebar{#1}}}

\newcommand{\la}{\ensuremath{\lambda}}

\newcommand{\be}{\ensuremath{\beta}}
\newcommand{\R}{\ensuremath{\mathbb{R}}}

\newcommand{\E}{\ensuremath{\mathbf{E}}}

\usepackage{mathtools}

\newcommand{\by}{\times}

\definecolor{light-gray}{gray}{0.80}

\definecolor{darkred}{rgb}{0.64, 0.0, 0.0}

\theoremstyle{definition}

\newtheorem*{thm*}{Theorem}

\DeclareMathOperator*{\argmin}{arg\,min}

\newcommand{\homework}[4]{
   \pagestyle{myheadings}
   \thispagestyle{plain}
   \newpage
   \setcounter{page}{1}
   \noindent
   \begin{center}
   \framebox{
      \vbox{\vspace{2mm}
    \hbox to 6.28in { {\bf CS 533:~Natural Language Processing \hfill {\small (#2)}} }
       \vspace{6mm}
       \hbox to 6.28in { {\Large \hfill #1  \hfill} }
       \vspace{6mm}
       \hbox to 6.28in {  {\it Instructor: {\rm #3}} \hfill}
      \vspace{2mm}}
   }
   \end{center}
   \vspace*{4mm}
}

\fancypagestyle{firstpage}{
  \fancyhf{} 
  \fancyfoot[L]{\small Essential AI (2025)} 

}

\title{Practical Efficiency of Muon for Pretraining}

\author{Essential AI \\
San Francisco, CA \\
\href{mailto:research@essential.ai}{\texttt{research@essential.ai}}}

\begin{document}
\pagestyle{plain} 
\maketitle
\thispagestyle{firstpage}

\begin{abstract}
We demonstrate that Muon, the simplest instantiation of a second-order optimizer, explicitly expands the Pareto frontier over AdamW on the compute-time tradeoff. 
We find that Muon is more effective than AdamW in retaining data efficiency at large batch sizes, far beyond the so-called critical batch size, while remaining computationally efficient, thus enabling more economical training. We study the combination of Muon and the maximal update parameterization (muP) for efficient hyperparameter transfer and present a simple telescoping algorithm that accounts for all sources of error in muP while introducing only a modest overhead in resources. We validate our findings through extensive experiments with model sizes up to four billion parameters and ablations on the data distribution and architecture.
\end{abstract}

\vspace{-0mm}
\begin{figure}[h]
\begin{center}
\begin{tabular}{cc}
\hspace{-10mm}\includegraphics[width=12cm, height=7cm]{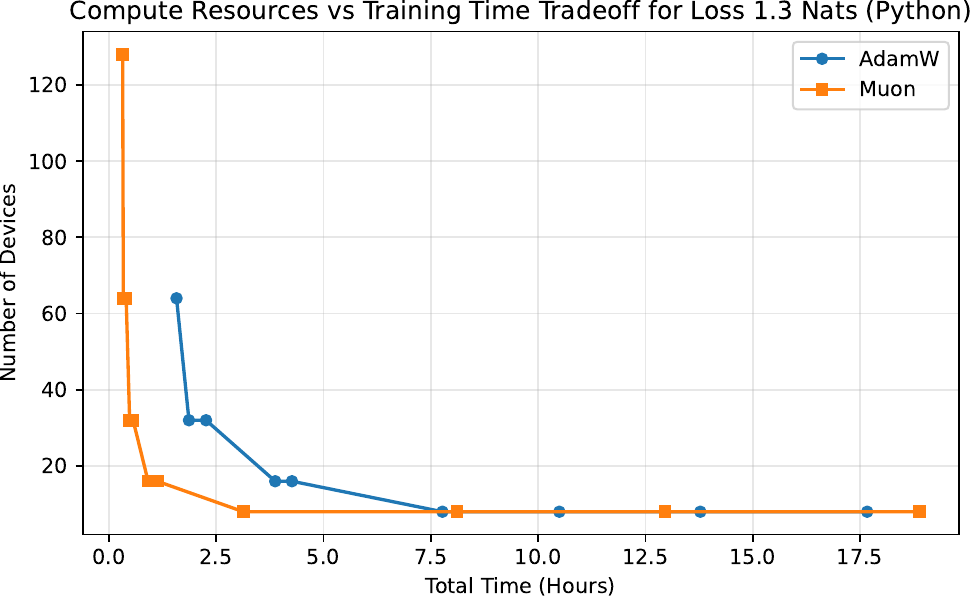}
\end{tabular}
\end{center}
\caption{Muon \citep{jordan2024muon} explicitly expands the Pareto frontier over AdamW on the compute-time tradeoff by retaining data efficiency at large batch sizes.}
\label{fig:abstract}
\end{figure}
\newpage

\section{Introduction}
\label{sec:introduction}

Adam \citep{kingma2014adam} and AdamW \citep{loshchilov2017decoupled} have empirically dominated the landscape of neural network optimizers in recent years. Superior generalization from the same training data (data-efficiency) at marginal increase in compute cost enabled them to gradually replace other first-order incumbents. By exploiting the statistical dependencies between model parameters, second-order optimizers have the potential to challenge the status quo. Recent second-order optimizers \citep{jordan2024muon,vyas2024soap} indeed promise data-efficiency over AdamW for pretraining, without compromising the overall computational advantage. However, these efforts rely on wall-clock time \citep{algoperf} or FLOPs to abstract away from algorithmic complexity and implementation details. But, this is insufficient to conclude their advantage over AdamW. Pretraining workloads come in various model shapes, data sizes, and resource affordances. To fully explain the practicality of an optimizer one needs to capture the fundamental tradeoff between compute and time resources---the ability to reduce total training time by using more devices. Since this trade-off is what makes large-scale pretraining feasible, it is crucial to characterize how it changes when comparing optimizers for pretraining. 

A key element in the compute-time tradeoff is the batch size. We may process data faster by increasing the batch size (i.e., by increasing the number of devices in distributed training) but likely at the cost of data efficiency - the loss reduction the model achieves per token. A ``good'' optimizer should therefore support large batch sizes to facilitate faster training while remaining persistently data-efficient. Simultaneously, it must remain computationally efficient by maintaining or improving FLOP efficiency and hardware utilization, ideally with any reasonable implementation. How can we take into account all these considerations to show that a second-order optimizer yields a more favorable compute-time tradeoff over AdamW?

To answer this question, we directly plot the compute and time resources required to achieve the same target loss using different optimizers. On this two-dimensional plane, optimizers are represented as iso-loss frontiers that marginalize over all other variables affecting the final tradeoff, thus admitting effective comparison. To study the tradeoff, we focus on Muon \citep{jordan2024muon}, the simplest instantiation of a second-order optimizer. We build on the recent work by \citet{liu2025muon}, which shows that Muon is more FLOP-efficient than AdamW, and make the stronger claim that Muon expands the Pareto frontier on the compute-time tradeoff with variable batch sizes, thereby strictly increasing the practitioner's flexibility in resource allocation (Figure~\ref{fig:abstract}). To analyze this behavior, we present a study of the relative data efficiency of Muon over AdamW in the large batch size regime. 

\begin{figure}
    \centering
    \includegraphics[width=.8\linewidth]{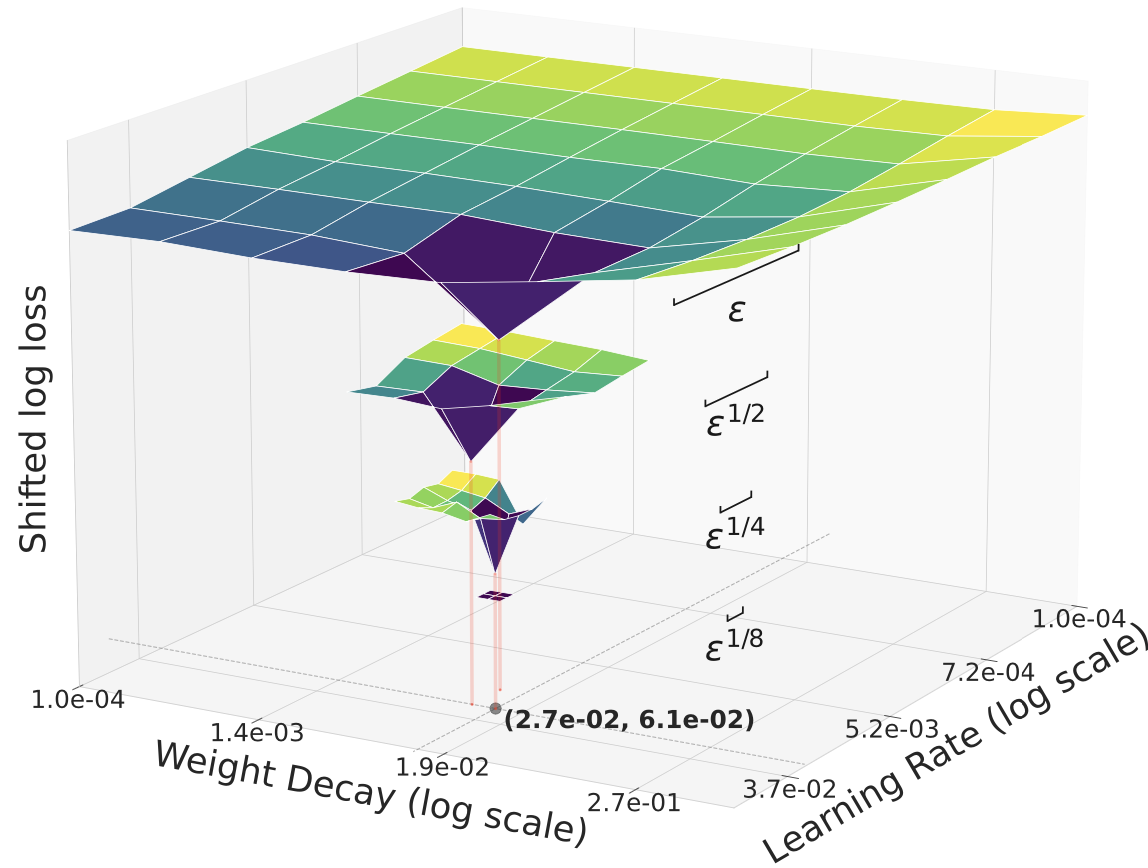}
    \caption{Telescoping algorithm applied to weight decay and learning rate. The loss for each model is shifted to a small offset and then presented on a logarithmic scale to exaggerate the minimum for visualization. The red vertical lines highlight the optimal hyperparameters for comparison. The narrowest model (smallest $n$) has the most grid points (the topmost surface) and the coarsest mesh with cell size $\epsilon = 1/(m-1)$ (see Eq.~\ref{eq:epsilon}), with $m$ being the number of sampled points for each hyperparameter. Submeshes at each subsequent doubling of width (lower surfaces) shrink in size, highlighting how the minima of each surface remain close across widths. Yellow regions indicate higher losses; color scales differ per stage for clarity. The shift in the red lines, marking the minima at each of the four layers, is proportional to $\alpha/n$ in Eq.~\ref{eq:opt-shift}. The largest grid is $8\times 8$, followed by $5\times 5$, $3\times 3$ and $1\times 1$, approximately following our geometric schedule in Alg.~\ref{alg:tele}. The third level, with cell size $\epsilon^{1/4}$ has extra points plotted to demonstrate that the true minimum is contained in the internal $3\times3$ subgrid. The final layer of the telescope, before training the final model, with cell size $\epsilon^{1/8}$ is $3\times 3$ rather than $1\times 1$ to demonstrate that the selected point is a true minimum (is locally flat).}
    \label{fig:telescope}
\end{figure}

In the second part of the paper, we address another core challenge in optimization for pretraining: how to choose optimal hyperparameters. We focus on the maximal update parameterization (muP) \citep{yang2022tensor}, a principled approach to hyperparameter transfer. It stipulates certain weight initialization and learning rate scaling such that optimal hyperparameters identified on a small model remain effective on a large one, reducing the burden of hyperparameter search at a large model scale. We demonstrate successful hyperparameter transfer with muP under Muon, extending known results under AdamW. While muP is asymptotically correct in the limit on the network width and search granularity, in practice it suffers from estimation errors and requires multi-scale tuning over different model sizes. We develop a simple ``telescoping'' algorithm for training a final model of cost $C$ and width $N$ that adjusts the search grid across model scales to account for these estimation errors with an additional modest $O(C\log{(N)})$ compute cost (Figure~\ref{fig:telescope}).

Our paper serves as a practitioner's guide on optimization for pretraining. Our final recommendation is to choose Muon over AdamW because it increases flexibility in resource allocation by remaining data-efficient with large batch sizes, and to combine it with muP for compute-efficient hyperparameter search at scale. We back our findings through extensive experiments, varying the model size up to 4 billion parameters and batch size up to 16 million tokens, ablating the effect of the data distribution and architecture.\footnote{All our experimental artifacts will be released at: \url{https://huggingface.co/EssentialAI}.}

\section{Muon Improves the Compute-Time Tradeoff}
\label{sec:optimizer}

\subsection{Review of Muon}
\label{sec:review-muon}

A transformer language model primarily consists of matrix weights $W \in \R^{m \by n}$ parameterizing the feedforward and attention layers. We assume $m \leq n$ without loss of generality. A weight update takes the form of $W_{t+1} = W_t - \eta_t O_t$ where $O_t \in \R^{m \by n}$ is some transformation of the gradient $G_t \in \R^{m \by n}$ on the $t$-th batch.  
Muon can be seen as matrix-structured steepest descent with spectral norm regularization:
\begin{align*}
    O_t = \argmin_{O \in \R^{m \by n}:\; \norm{O}_2 \leq 1} \tr{G_t^\top O}
\end{align*}
It is easy to show that $O_t = U V^\top$ is an optimal solution where $G_t = U \Sigma V^\top$ is a singular value decomposition (SVD) \citep{bernstein2024old}. Muon avoids explicit computation of SVD by Newton-Schulz iteration \citep{kovarik1970some,bjorck1971iterative}. In practice, we combine Muon with Nesterov momentum, learning rate scaling, and (coupled) weight decay. Thus the only state Muon maintains is the first moment $M_t \in \R^{m \by n}$ initialized to zero. 
In the $t$-th update, given the gradient $G_t$ we compute
\begin{align}
    M_t &= G_t + \be M_{t-1} \notag \\
    O_t &= \textbf{NewtonSchulz}(G_t + \be M_t) \notag \\
    W_{t+1} &= W_t - \eta_t ((0.2\sqrt{n}) O_t + \la W_t) \label{eq:muon-step}
\end{align}
where $\be, \la \in \R$ are momentum and weight decay hyperparameters. The constant $0.2 \sqrt{n}$, proposed by  
\citet{liu2025muon}, normalizes the Muon update to have a similar RMS value as the AdamW update.\footnote{0.2 is the empirically observed RMS of a typical AdamW update. $1/\sqrt{n}$ is the RMS of a perfectly normalized $O_t$, which follows immediately from the fact that the squared Frobenius norm is the sum of squared singular values.
}
This ensures that the same learning rate schedule and weight decay work well for both Muon and AdamW, which is useful because AdamW is used to update embedding and normalization parameters. \citet{jordan2024muon} provide well-tuned hyperparameters for Newton-Schulz that rapidly flatten all singular values to range (0.7, 1.3).

\subsubsection{Why we focus on Muon}

While Muon is not the only viable second-order optimizer, we limit our scope to Muon for the following reasons. 
First, it has the simplest derivation and implementation, allowing us to focus on characterizing the practical benefits of second-order methods. Second, we can reduce more sophisticated algorithms such as Shampoo \citep{gupta2018shampoo, anil2020scalable} and Soap \citep{vyas2024soap} directly to Muon under simplifying assumptions (Appendix~\ref{app:muon-reduction}). Thus Muon is in some sense the minimal version of this class of optimizers.
Third, Muon has the lightest memory footprint of all the optimizers we considered (even lighter than AdamW) since it only maintains the first moment, reducing the infrastructure work needed for distributed training. Fourth, the FLOP overhead of Muon over AdamW diminishes in the batch size $B$ as $\Theta(\frac{m}{B})$, making Muon a promising candidate for large-batch training. Fifth, given the default hyperparameters in Newton-Schulz iteration, Muon only needs tuning over the maximum learning rate and weight decay, which are furthermore compatible with AdamW under the constant  $0.2 \sqrt{n}$ rescaling (\cite{liu2025muon}) and make comparison with AdamW especially convenient. 

\subsection{Experimental Setup}
\label{sec:setup}

\paragraph{Architectures.} We use a modern decoder-only transformer model based on Gemma 3 \citep{team2025gemma}, which replaces the soft-capping method in Gemma 2 with QK-norm for attention stability. Unlike Gemma 3 which interleaves local and global attention layers, we only use global attention layers for simplicity. 
We compare Gemma 3 with Gemma 2 in our ablation studies.

\paragraph{Data distributions.} A typical pretraining data mix consists of many different sources (GitHub code, math blogs, books, arXiv articles, general web, etc.). We use data distributions with markedly different compression ratios and target capabilities, to reflect the heterogeneity of the pretraining mix.
Specifically, we focus on text and code.
For text, we use high-quality general web text from DCLM \citep{li2024datacomp}.
For code, we use Python code obtained from our in-house filtered version of the Stack V2 \citep{lozhkov2024starcoder}.
We chunked input sequences to length 8192 using the Llama3 tokenizer with the vocab size of 128K.

\paragraph{Token budget.}
To ensure training is converged, we make an effort to use enough tokens to meet or exceed the Chinchilla-optimal token budget under our resource constraints \citep{hoffmann2022training}. Specifically, we trained the 100M model size on 10B tokens ($5\times$), 500M on 25B tokens ($2\times$), and each of the 1B, 2B, and 4B model sizes on 50B tokens ($2.5\times$, $1.25\times$, and $0.625\times$).

\paragraph{Optimizer implementations.} We use the AdamW implementation (\texttt{optax.adamw}) in the Optax library. We use a straightforward in-house Jax implementation of Muon described in Section~\ref{sec:review-muon}, applying \eqref{eq:muon-step} to all layers except embedding and normalization (Appendix~\ref
{app:muon-imp}). For the latter layers, we use Optax Adam with the same learning rate and weight decay while holding $\be_1 = \be_2 = 0.95$ fixed. We train our models on TPU v5p chips, achieving close to 50\% model FLOP utilization (MFU).

\paragraph{Hyperparameter tuning.} We ran a fine-grained sweep at the 100M model scale, then validated the efficacy of the chosen hyperparameters at the 500M scale. For the 1B model, we ran a smaller sweep across different batch sizes, and then carried those findings to the 2B and 4B experiments. In all runs, we use the learning rate schedule of a linear warmup and cosine decay to 0.1 of the max learning rate.

\paragraph{Initial verification of Muon's performance}
We were able to quickly obtain positive results with the above setup (Appendix~\ref{app:initial-verification}). Specifically, we confirmed that at any given number of steps, Muon consistently achieved a lower training loss.\footnote{Since we do not epoch over data, we use the training loss on a new batch as a proxy for generalization error for convenience. While this is limited in quantity compared to using a separate held-out dataset, it achieves the same effect when averaged over many batches.} We also confirmed that Muon achieves a target loss faster than AdamW in wall time.

\begin{figure}[t]
\begin{center}
\begin{tabular}{ccc}
\hspace{-4mm}\includegraphics[width=4.5cm]{karl_images/tradeoff_py_1.3.pdf}
&
\hspace{-3mm}\includegraphics[width=4.5cm]{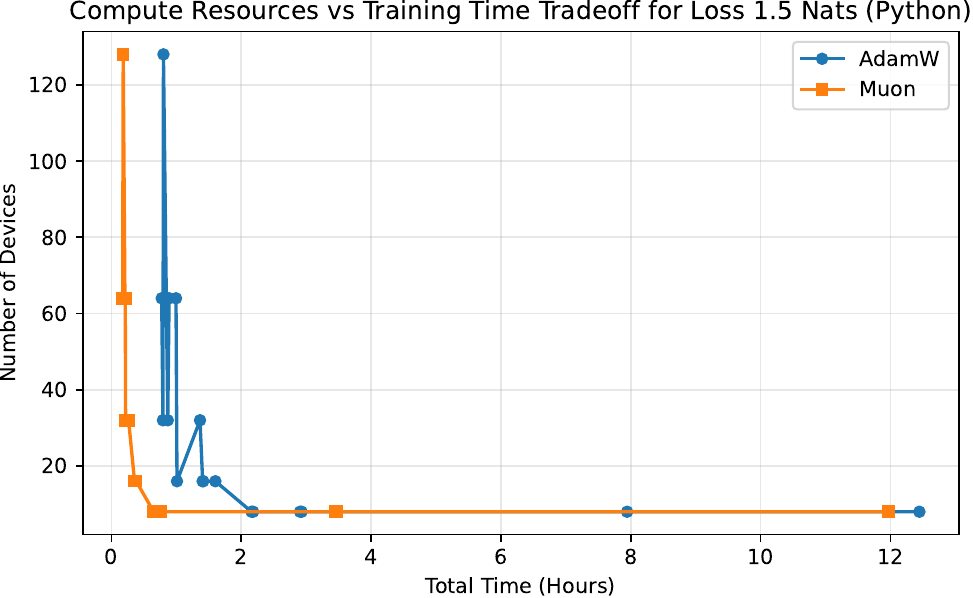}
& 
\hspace{-3mm}\includegraphics[width=4.5cm]{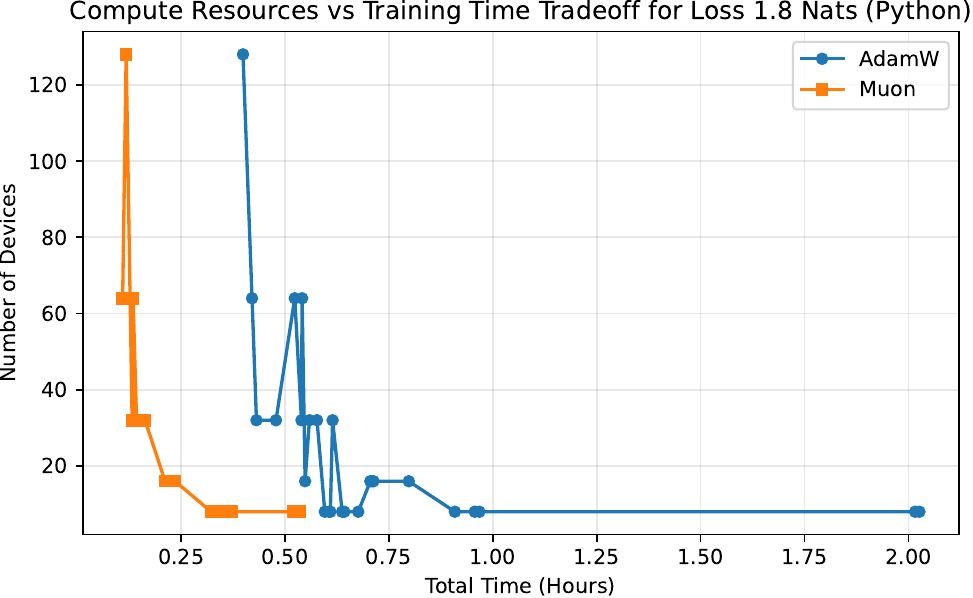} \\
\hspace{-4mm}\includegraphics[width=4.5cm]{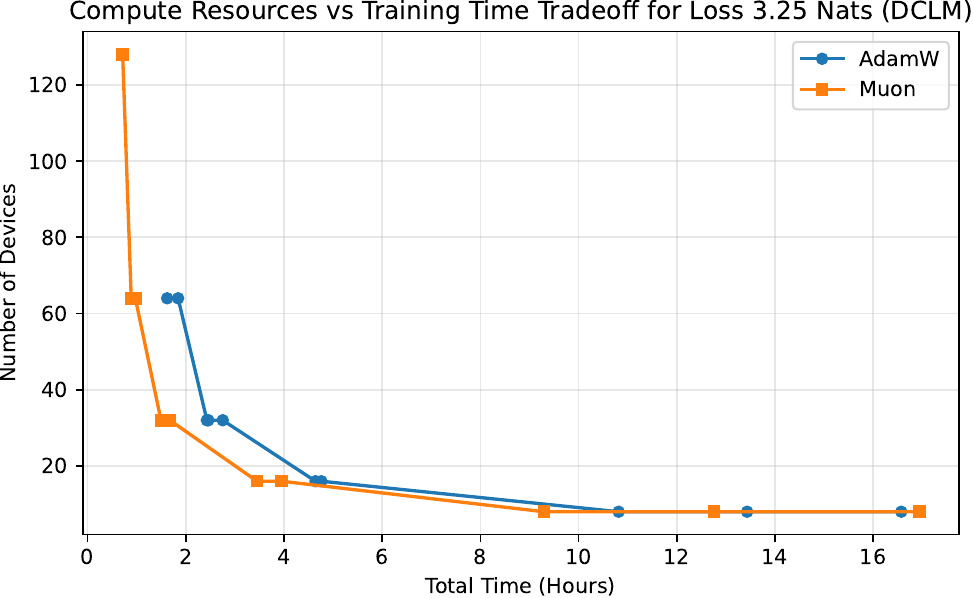}
&
\hspace{-3mm}\includegraphics[width=4.5cm]{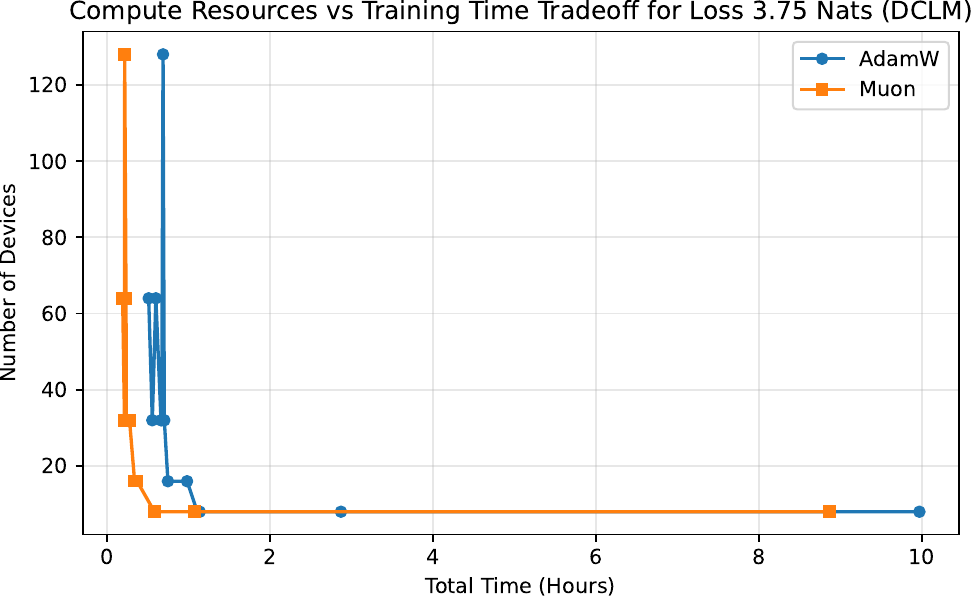}
& 
\hspace{-3mm}\includegraphics[width=4.5cm]{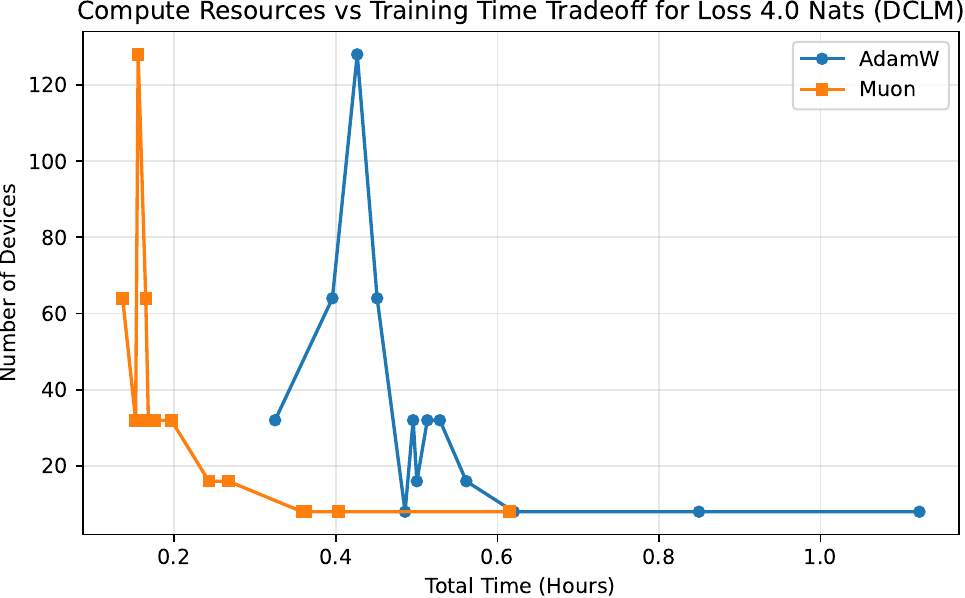}
\end{tabular}
\end{center}
\caption{Muon expands the Pareto frontier over AdamW on the compute-time tradeoff at various loss thresholds on Python (top) and DCLM (bottom).}
\label{fig:tradeoff-main}
\end{figure}

\subsection{The Compute-Time Tradeoff for Pretraining}

Conventional approaches that simply compare optimizers in wall time or FLOP efficiency do so at fixed resources (i.e., devices) and fail to characterize the tradeoff between resources and training time.
Instead, we compare optimizers as iso-loss curves on the compute-time plane, by measuring the total training time to reach a target loss as a function of the number of devices and correspondingly the batch size. The gap between the curves represents novel economically feasible options (e.g., shorter run or fewer devices than what is possible under the other optimizer).

We simulate how a practitioner may use more devices to reduce the total training time by training 500M parameter models over 13 batch sizes ranging from 128K to 16M with data parallelism, where we vary the number of devices from 8 to 128 in a realistic way to ensure enough on-device memory.\footnote{Specifically, we use 8 devices for batch size under 1M, 16 for under 2M, 32 for under 4M, 64 for under 8M, and 128 for above 8M. } 
We show the resulting plots in Figure~\ref{fig:tradeoff-main}.
The plot shows that Muon explicitly 
expands the Pareto frontier on the compute-time tradeoff, thereby strictly increasing
the practitioner’s flexibility in resource allocation. 
The expansion is consistent on different thresholds and data distributions. See Appendix~\ref{app:tradeoff-plots} for overlaid thresholds.

The plot gives a concise summary of Muon's practical efficiency over AdamW, marginalizing over all variables that affect the final compute-time tradeoff in complex and possibly competing ways. For instance, increasing the batch size may increase the per-device MFU (assuming enough devices) and thereby increase the number of FLOP/s, but it may simultaneously decrease the data efficiency of the optimizer and thereby require more FLOPs to reach the same loss.
The plot also abstracts away implementation details of the optimizers, which may raise the concern that the gain unfairly comes from AdamW's suboptimal implementation. However, we use the standard Optax AdamW implementation which we view as close to optimal. On the other hand, our Muon implementation is quite naive and thus likely to only strengthen the efficiency gain with further optimization.

\subsection{Muon's Relative Data Efficiency Over AdamW} 
\label{sec:data-efficieny}

As we add more devices with data parallelism, we need to increase the batch size correspondingly to ensure that the devices are not memory-bound. At the same time, the data efficiency of an optimizer begins to diminish once the batch size passes a certain threshold, requiring more tokens to reach the same loss. Thus we hypothesize that Muon improves the compute-time tradeoff by remaining more data-efficient than AdamW at large batch sizes.

\begin{figure}[t]
\begin{center}
\begin{tabular}{ccc}
\hspace{-4mm}\includegraphics[width=4.5cm]{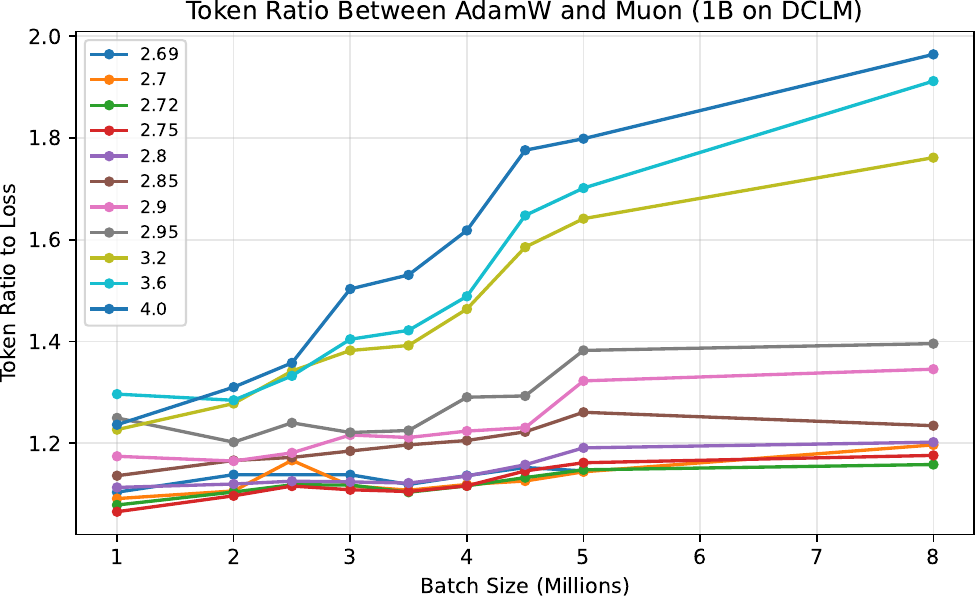}
& 
\hspace{-2mm}\includegraphics[width=4.5cm]{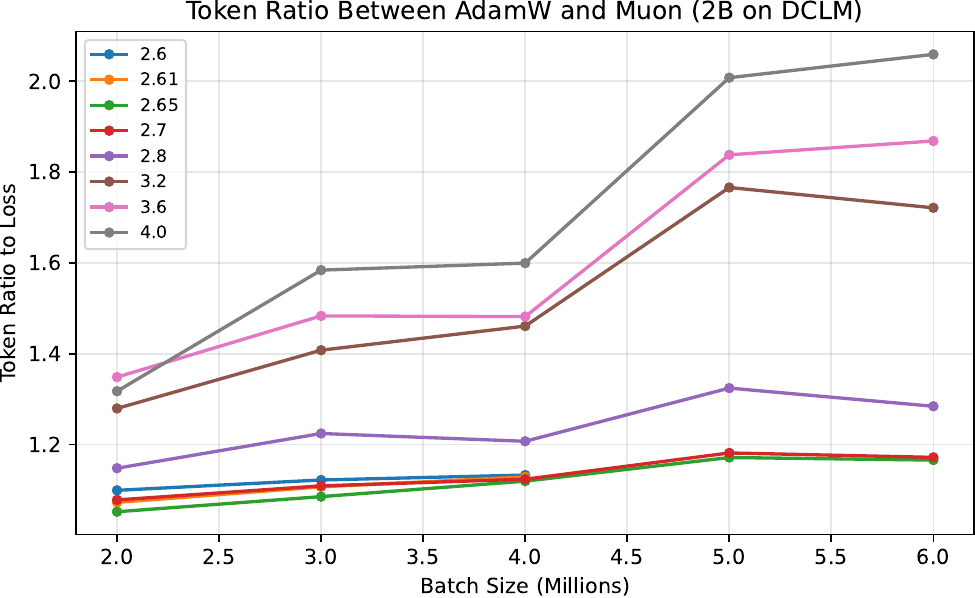}
& 
\hspace{-2mm}\includegraphics[width=4.5cm]{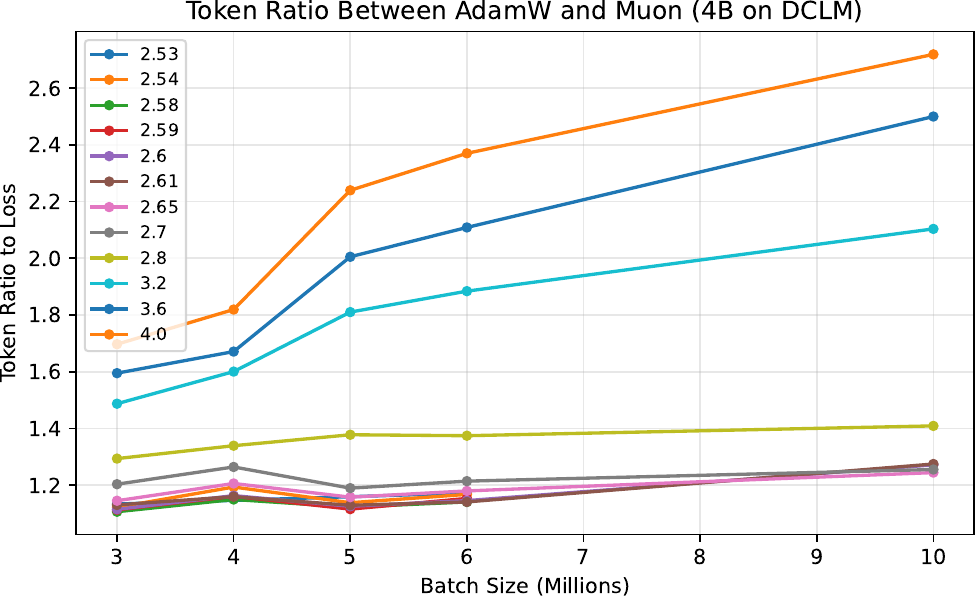} \\
\hspace{-4mm}\includegraphics[width=4.5cm]{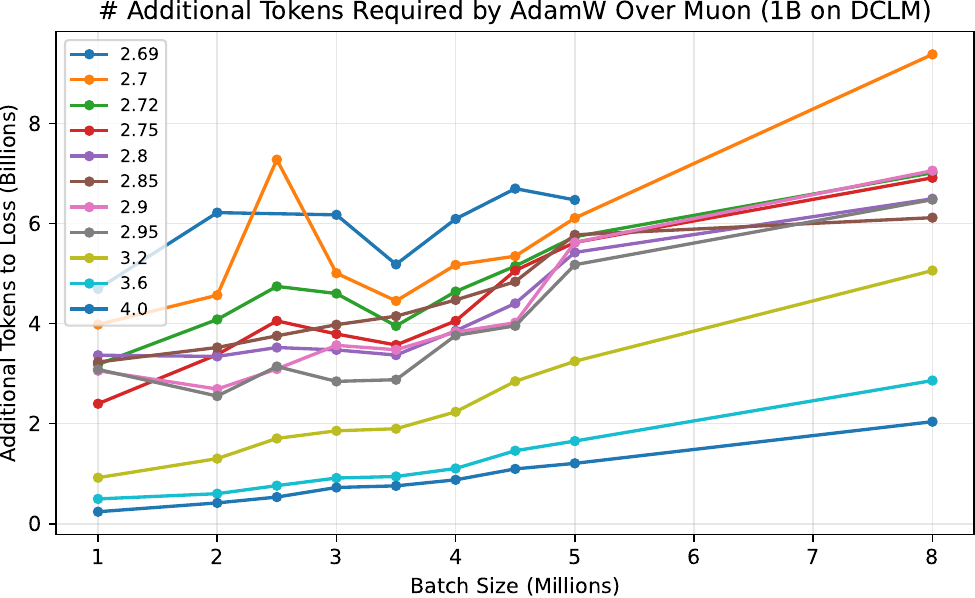}
& 
\hspace{-2mm}\includegraphics[width=4.5cm]{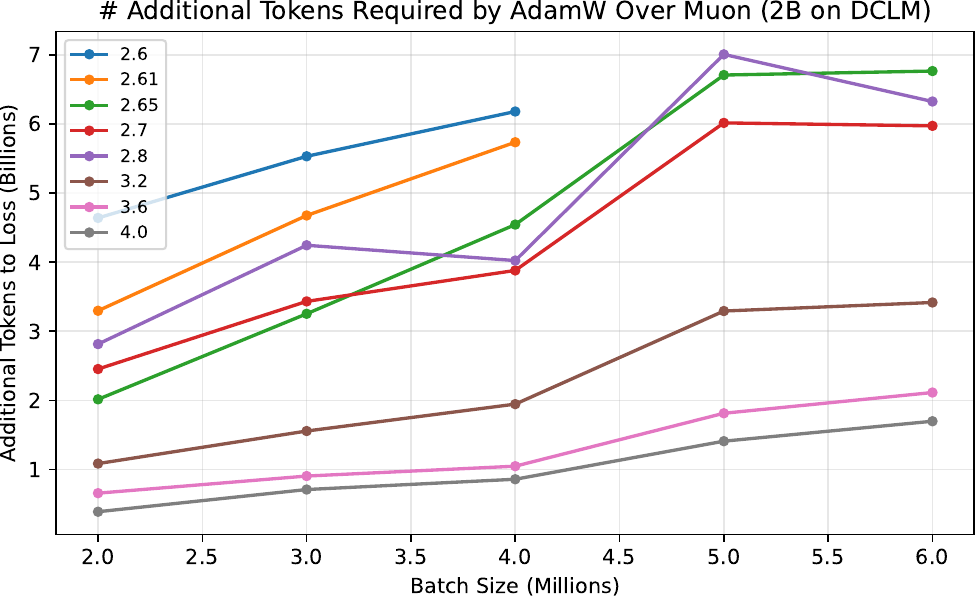}
& 
\hspace{-2mm}\includegraphics[width=4.5cm]{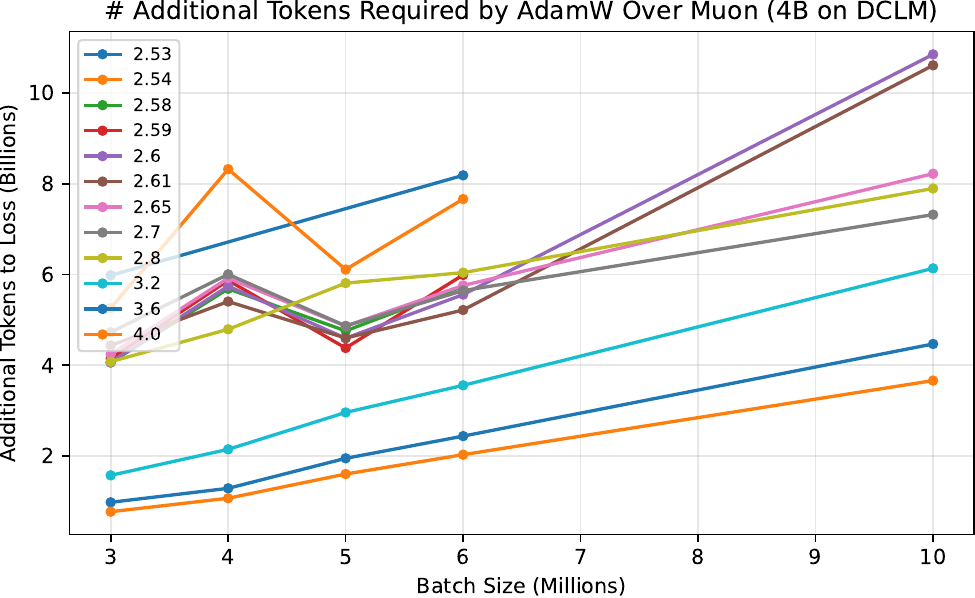} \\
\hspace{-4mm}\includegraphics[width=4.5cm]{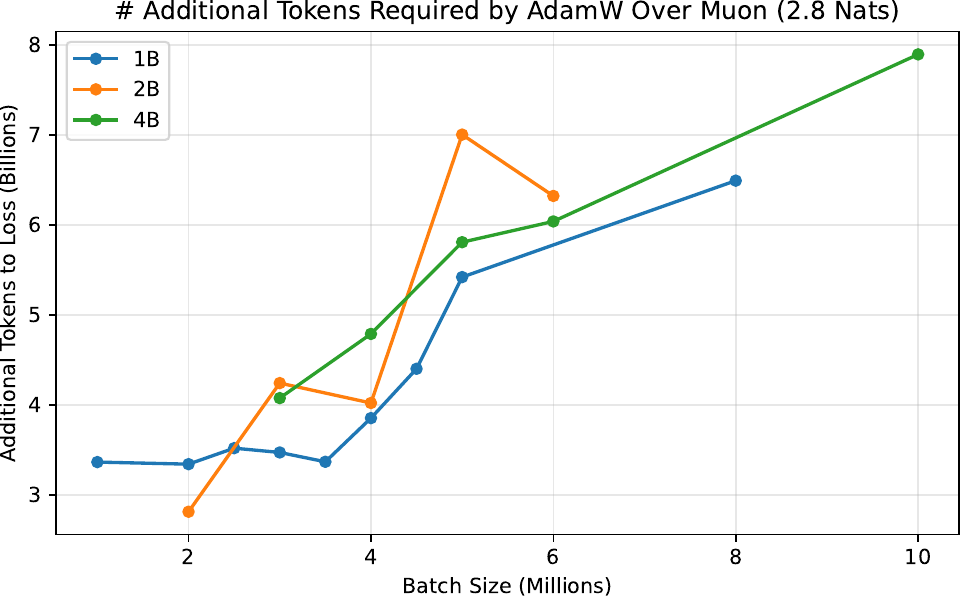}
& 
\hspace{-2mm}\includegraphics[width=4.5cm]{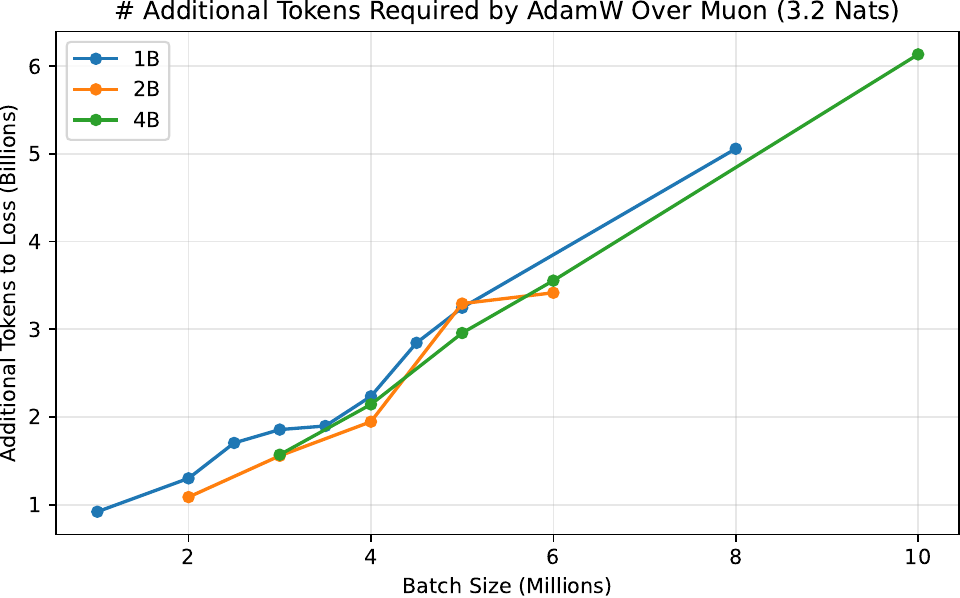}
& 
\hspace{-2mm}\includegraphics[width=4.5cm]{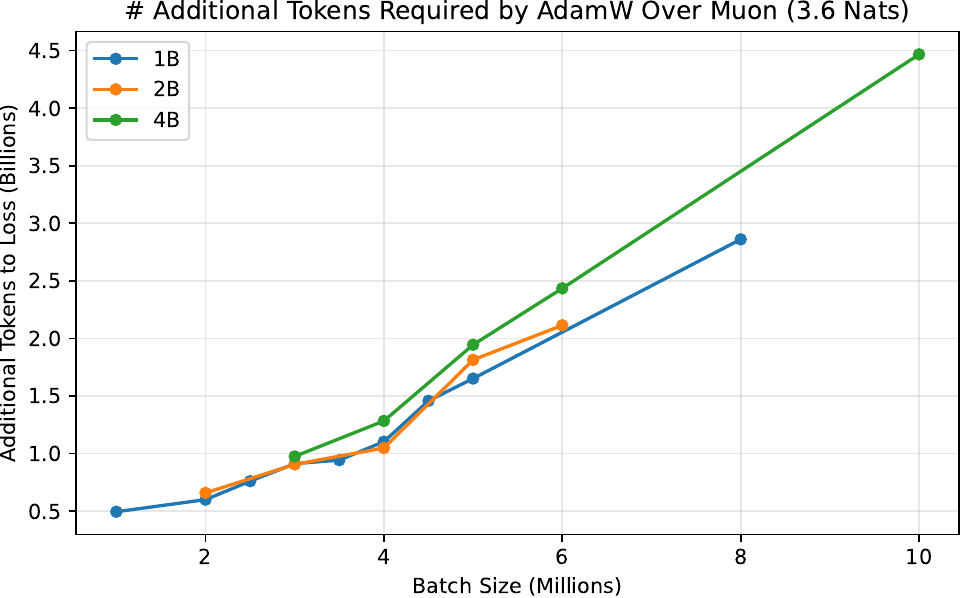} 
\end{tabular}
\end{center}
\caption{Relative data efficiency of Muon over AdamW. (Top) Token ratio vs batch size plots for different target losses across model sizes. (Middle) Corresponding token difference plots. (Bottom) Token difference across model sizes at fixed target losses}
\label{fig:token-ratio}
\end{figure}

To characterize the relative data efficiency of Muon over AdamW, we propose measuring the ratio of their token consumptions:
\begin{align}
    R_L(B) = \frac{T_{L,A}(B)}{T_{L,M}(B)} = 1 + \frac{T_{L,A}(B) - T_{L,M}(B)}{T_{L,M}(B)}  \label{eq:token-ratio}
\end{align}
where $T_{L,A}(B)$ is the number of tokens consumed to reach a target loss $L$ at batch size $B$ under AdamW, similarly $T_{L,M}(B)$ for Muon.
The ratio represents data overhead: how many more tokens does AdamW need to reach the same loss, relative to Muon's own token consumption? If $R_L(B)$ remains above $1$, and is increasing or at least nondecreasing when $B$ is large, the relative data efficiency of Muon over AdamW does not vanish even in the over-large batch size regime.\footnote{We note that $R_L(B)$ translates to non-optimizer FLOP overhead in training, assuming the standard $6dT$ FLOP count approximation for training a transformer language model with $d$ parameters on $T$ tokens.}

Figure~\ref{fig:token-ratio} (top row) shows empirical observations with 1B model size on DCLM. 
We see that the token ratio is above 1 and generally increasing at higher losses and at least constant at lower losses, suggesting that Muon's relative batch size benefit persists beyond any fixed batch sizes. 
It also implies that the unnormalized additional number of tokens $T_{L,A}(B) - T_{L,M}(B)$ strictly increases in $B$ (middle row). We observe that the number of extra tokens required by a model is surprisingly agnostic to model size (bottom row). For instance, to reach a target loss of 3.2 nats, all three of the 1B, 2B and 4B model need around 2B extra tokens with AdamW than Muon.

The empirical fact that $R_L(B) > 1$ is nondecreasing means the non-optimizer FLOP overhead of AdamW stays constant even at large batch sizes. This gives Muon an edge in trading off compute resources (i.e., more devices with bigger batch sizes) and training time, despite the fact that Muon performs more FLOPs than AdamW.

\subsection{Connection to Critical Batch Size}
\label{sec:batch-size}

\vspace{-0mm}
\begin{figure}[t]
\begin{center}
\begin{tabular}{cc}
\hspace{-1mm}\includegraphics[width=6.9cm]{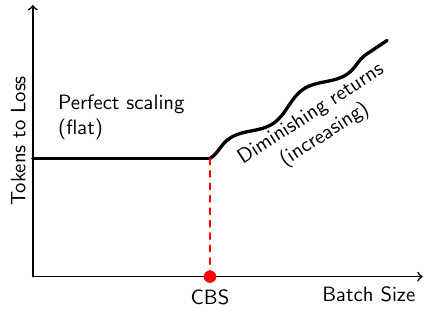}
& 
\hspace{-4mm}\includegraphics[width=6.5cm]{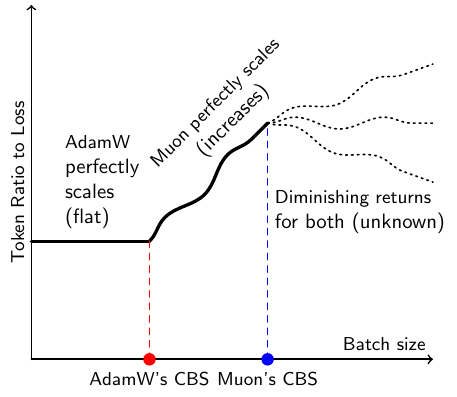}
\end{tabular}
\end{center}
\caption{\emph{Token ratios to loss} give a clearer picture of the practical advantages of Muon over AdamW, compared to \emph{tokens to loss}}
\label{fig:critical-figures}
\end{figure}

A great deal of research on the impact of batch size on training focuses on the idea of ``critical'' batch size, which is broadly defined as the largest batch size that allows for almost linear or ``perfect'' scaling (e.g., doubling the batch size halves the number of steps) beyond which we receive diminishing returns \citep[e.g.][]{zhang2019algorithmic}.
We can relate the token ratio $R_L(B)$ to the critical batch size as follows.
We can define the critical batch size as a value $B_L^\star$ such that (1) the number of tokens required to reach the loss remains constant $T_L(B) =  T_L^\star$  for batch sizes $B \leq B_L^\star$, and (2) $T_L(B) = \mathrm{inc}_L(B)$ is some increasing function for $B > B_L^\star$. 
Even if Muon's critical batch size is larger than Adam's, $R_L(B)$ may either increase, decrease, or stay constant depending on the tail behavior of token consumption under the two optimizers (Figure~\ref{fig:critical-figures}).
Our empirical finding is that $R_L(B)$ is nondecreasing (Figure~\ref{fig:token-ratio}), providing information on the post-critical batch size regime.
Appendix~\ref{app:critical-ours} gives more detailed analysis.
\section{Choosing Hyperparameters for Muon}
\label{sec:hyper}

Up until this point, we have discussed the benefits of replacing AdamW with Muon in pretraining. To utilize an optimizer efficiently, it is also important to be able to calibrate it efficiently---specifically to be able to select near-optimal hyperparameters. \cite{moonshot2025tweet} and \cite{liu2025muon} leave open the question of the compatibility of Muon with standard hyperparameter transfer techniques, such as muP. Hence, after introducing the sources of error in muP in Sec.~\ref{sec:sources}, in Sec.~\ref{sec:mup-experiments} we give the first empirical demonstration of muP used to calibrate hyperparameters for a large language model using Muon as the optimizer. We note that while \cite{wang2024set} considers the hyperparameter transfer of decoupled weight decay, we consider the coupled setting and hence inherit the transfer properties discussed in \cite{yang2022tensor}. First, in the next section, we introduce the high-level ideas of muP. 

\subsection{The Maximal Update Parameterization (muP)}
\label{sec:mup}

Brute-force grid searches over hyperparameters become intractable for large models. For $k$ hyperparameters each on an $m$-point grid, we must train $m^k$ separate models, which can be prohibitively expensive. As an example, the largest model trained in this section has $3.7\text{B}$ parameters and took $100$ hours on $128$ TPU v5ps to process $160\text{B}$ tokens; a modest $8\times 8$ grid search on the full model would take nearly $820{,}000$ device hours. Comparatively, \cite{liu2024deepseek} trained a $671\text{B}$ model with $37\text{B}$ active parameters with $14.8\text{T}$ tokens and required only $2.788\text{M}$ H800 GPU hours, so such a grid search would consume nearly 30\% of the total compute budget while training a substantially smaller model.

Hyperparameter transfer is a practical alternative: train a smaller “proxy” model, tune its hyperparameters, and apply the resulting settings to the larger model. However, without further insight, there is no guarantee that the optimal hyperparameters for the smaller model work well on the larger one. MuP \citep{yang2022tensor,yaida2022meta} solves exactly this issue, ensuring consistent and predictive hyperparameter transfer across model scales. We introduce the core motivating discussion in App.~\ref{app:mup} and provide explicitly our choice of initialization, multiplier and learning rate scaling in Table~\ref{table:params}.

Specifically, we define a width-dependent parameterization of the network using three width-dependent functions $a(n), b(n)$ and $c(n)$. Each weight matrix $W_\ell$ at layer $\ell$ is rescaled as $W_{\ell} = a(n) w_{\ell}$, where $w_{\ell}$ are the trainable parameters, each entry of $w_{\ell}$ is initialized as $w_{\ell} \sim \mathcal{N}(0, b(n))$ and the learning rate is set to $c(n)\eta_0$ for some width-independent base rate $\eta_0$.

\begin{figure}[H]
\centering
\renewcommand{\arraystretch}{1.5}
\begin{tabular}{lccc}
\toprule
 & \textbf{Input Weights \& Biases} & \textbf{Output Weights} & \textbf{Hidden Weights} \\
\midrule
\textbf{Multiplier} $a(n)$
& $\sqrt{\mathrm{fan\_out}}$
& $\tfrac{1}{\sqrt{\mathrm{fan\_in}}}$
& $1$ \\
\textbf{Initialization\ Variance} $b(n)$
& $\tfrac{1}{\mathrm{fan\_out}}$ 
& $\tfrac{1}{\mathrm{fan\_in}}$ 
& $\tfrac{1}{\mathrm{fan\_in}}$ \\
\textbf{Learning Rate} $c(n)$
& $\tfrac{1}{\sqrt{\mathrm{fan\_out}}}$ 
& $\tfrac{1}{\sqrt{\mathrm{fan\_in}}}$
& $\tfrac{1}{\mathrm{fan\_in}}$ \\
\bottomrule
\end{tabular}
\caption{muP hyperparameter scaling rules, reproduced from \cite{yang2022tensor}. 
Here, $\mathrm{fan\_in}$ and $\mathrm{fan\_out}$ denote the input and output dimensions, respectively, for a given weight matrix, e.g. for the weight matrix in the $\ell^{\text{th}}$ layer, $W_\ell \in \mathbb{R}^{m\times n}$ $\mathrm{fan\_in}$ is $n$ and $\mathrm{fan\_out}$ is $m$.}
\label{table:params}
\end{figure}

\subsection{Sources of Error}
\label{sec:sources}
This section analyzes the dominant sources of error in hyperparameter transfer under muP, introduced in the previous section. In the next section (Sec.~\ref{sec:telescope}), we will introduce a simple algorithm for controlling and suppressing the errors discussed in this section. The core analysis we present, depends on the well-known fact that neural networks often admit well-defined infinite width limits (e.g. \cite{hanin2023random}). That is, given a large language model with width parameters (e.g. number of heads, hidden dimension and MLP dimension) there is a family of large language models $f(x, n)$ for hyperparameter $x$, that we can construct by increasing the width $n$. The core assumption we will make is that this limit exists, call it $f_0(x)$, and is smooth in $n$, the network width. Choosing a simple re-parameterizization of the network in $1/n$ (which is small, for large $n$) instead of $n$, we can apply a Taylor expansion to $f(x, 1/n)$ as

\begin{equation}\label{eq:nnfunc}
    f(x, 1/n) = f_0(x) + \frac{f_1(x)}{n} + O\paren{\frac{1}{n^2}}.
\end{equation}

(The full analysis is more complicated, as discussed in \cite{yang2019scaling} and \cite{yaida2022meta}, and in general we need to ensure that the infinite series can be truncated at finite order. For our setting, however, this is sufficient.) From this analysis alone, it is clear that there is imprecision introduced in using any finite width, $n$, for hyperparameter transfer. The theory introduced in \cite{yang2019scaling} and experimentally verified in \cite{yang2022tensor} enables principled transfer of hyperparameters from small to large models by aligning their training dynamics across different model widths. However, numerous recent experiments (e.g., in the appendix of \cite{everett2024scaling}) demonstrate that such transfer is only approximately valid even at relatively large widths, with errors that diminish as the model becomes wider. This observation aligns with the central tenet of muP: hyperparameter transfer becomes exact only in the infinite-width limit. 

We identify two primary sources of error in the muP hyperparameter transfer protocol. First, for a loss function $\mathcal{L}$, the optimizing hyperparameter \( x^*(n) \) of $\mathcal{L}(f(x, n))$ at width \( n \) deviates from the optimizing hyperparameter of its infinite width counterpart \( x^* \) that optimizes $\mathcal{L}(f_0(x))$ due to \(1/n\)-order corrections. It follows from Eq.~\ref{eq:nnfunc} that the optimal hyperparameter $x^*(n)$ minimizing the loss shifts as
\begin{equation}\label{eq:opt-shift}
   x^*(n) = x^* - \frac{\alpha}{n} + O\paren{\frac{1}{n^2}},
\end{equation}
for some constant \(\alpha\) depending on the network and loss function (see App. ~\ref{sec:taylor} for the specific details). This finite-width bias represents an unavoidable source of error unless the proxy model is sufficiently large.

Second, even for stationary loss minima (i.e. $x^*(n) = x^*$), mesh-based approximations during the hyperparameter sweep introduce sampling error. If \(\hat{x}^*(n)\) is the discrete optimum estimated by evaluating a grid of hyperparameters on a model of width \(n\), then 
\begin{equation}\label{eq:epsilon}
\hat{x}^*(n) \approx x^*(n) + \varepsilon,
\end{equation}
where \(\varepsilon\) reflects the resolution of the sweep. In practice, coarse meshes, limited budget, or poorly chosen sweep ranges may lead to \(\varepsilon\) large enough to obscure critical features of the loss landscape. In the next section, we introduce an algorithm to control for both sources of error independently, with nearly optimal complexity.

\subsection{The ``Telescoping'' Protocol}\label{sec:telescope}
Algorithm~\ref{alg:tele} presents a practical method for systematically controlling these errors. The intution behind our ``telescoping'' algorithm is that because we have the sources of error mentioned in Sec.~\ref{sec:sources}, we train models at a range of widths, but also reduce the number of grid points at the larger widths due to their increased cost. The idea of intelligently constraining and allocating resources for hyperparameter search is not new \citep{fetterman2023tune, li2018hyperband}, and our idea of logarithmically reducing the search space more generally fits in the category of hierarchical Bayesian optimization \citep{kennedy2000predicting}.

The accompanying parameterized analysis (Fig.~\ref{fig:telescope-analysis} in the Appendix) shows that, depending on the desired confidence level, the compute budget for the final model training may vary between $20\%$ and $99\%$ of the total training and hyperparameter tuning cost; the compute saved when compared to a brute force grid search is typically higher than $50\%$. As a result, the total savings in large-scale training remains substantial.

\begin{algorithm}[H]
\caption{Telescoping Algorithm for Hierarchical Hyperparameter Transfer}
\label{alg:tele}
\KwIn{Base model width $N_0$, final calibration width $N_c$, final model width $N$, number of hyperparameters $k$, number of sweep points $m$}
\KwOut{Sequence of optimal hyperparameters with controlled drift up to width $N$}

\textbf{Initialize:} Determine mesh size using a curvature estimate near the optimum at $N_0$. Set $n \gets N_0$\;

Perform hyperparameter grid sweep at width $N_0$ with full mesh\;

\While{$n < N_c$}{
    $n \gets 2n$ \tcp*[r]{Double the model width (Increases cost by 4$\times$)}
    Reduce number of sweep points per hyperparameter by factor $4^{-1/k}$\;
    Adjust mesh resolution accordingly (submesh of size $4^{-1/k}$)\;
    Perform the grid sweep over the refined mesh\;
}
\end{algorithm}

The core assumption behind the telescoping approach presented in Alg.~\ref{alg:tele} is that the optimal hyperparameter varies smoothly with model width, with a shift of order $1/n$. Under this assumption, when the width doubles (which increases the FLOPs by roughly a factor of four),  halving the mesh spacing suffices to track the $1/n$ drift while keeping the total number of samples of each hyperparameter at each stage proportional to $4^{-1/k}$. Thus, the cost of each successive stage is roughly constant. Crucially, this prevents \emph{under}-refinement (missing the shifted minimum) and \emph{over}-refinement (unnecessary fine-grained sweeps).

A sufficiently broad initial sweep at the smallest width $N_0$ usually leverages prior empirical knowledge about feasible hyperparameter ranges. As $n$ doubles, the new sweep narrows the mesh just enough to capture any incremental drift of the optimum. Absent any new peaks emerging at later widths—which empirical practice suggests is relatively rare—this ``half-meshing'' strategy is near-optimal. It balances the competing needs of accuracy in locating the optimal hyperparameter and computational efficiency in the sweep. Because each refinement stage maintains approximately constant cost (training is more expensive but the search space is reduced), we see that for a final model compute cost of $C$ and width $N$ this procedure only introduces an additional factor of $O(C\log({N}))$ to the computational requirements of hyperparameter tuning, with an optimum which is guaranteed to remain close to the true minimum at the largest model size. Hence, the final training run still consumes a significant fraction of the overall budget and the total tuning overhead remains modest. Ultimately, this telescoping scheme extends the standard muP\;``tune-up'' procedure by ensuring that errors are governed by the largest calibration width $N_c$ (which can be chosen to be $O(N)$), as opposed to the smallest width $N_0$, while incurring only an overhead of $O(C\log{(N)})$.

\subsection{Experiments}
\label{sec:mup-experiments}
We validate our telescoping algorithm on a family of transformer models aimed at a final size of approximately $3.7$B parameters. All models use a sequence length of $8192$, a batch size of $2^{23}\approx 8$ million tokens and a depth of $34$ layers. The smallest model in the family has head dimension $12$, $1$ key/value head, $2$ query heads, a base embedding dimension of $256$ and a base MLP dimension of $1024$. We do not scale the AdamW \(\epsilon\) parameter, following \cite{yang2022tensor} and \cite{everett2024scaling}, as our models are sufficiently small. Additionally, although \cite{yang2019scaling} demonstrates approximate hyperparameter transfer across model dimensions besides width, we note that related theoretical work (\cite{large2024scalable}) demonstrates that precise transfer along these dimensions requires architectural changes. Hence, we fix all other ``scale'' dimensions --- depth, sequence length, batch size and training steps --- and only vary width.

For the largest model (about $3.7\text{B}$ parameters), we train for $20,000$ steps, corresponding to $160$B tokens---approximately $2.2\times$ the Chinchilla-optimal budget ($72$B tokens) (\cite{hoffmann2022training}). Though this exceeds the nominal optimal budget, over-training can be beneficial when seeking a smaller inference footprint, often with negligible impact on final loss (\cite{mcleish2025gemstones}). The full hyperparameter sets are the point-wise products of $\mathcal{H}_q = \{2, 4, 8, 16, 20\}$, $\mathcal{H}_{kv} = \{1, 2, 4, 8, 10\}$, $\mathcal{D}_{\text{emb}} = \{256, 512, 1024, 2048, 2560\}$, $\mathcal{D}_{\text{mlp}} = \{1024, 2048, 4096, 8192, 10240\}$, $\mathcal{S} = \{0.07\text{B}, 0.20\text{B}, 0.67\text{B}, 2.40\text{B}, 3.67\text{B}\}$ with $\mathcal{H}_q$ the number of query heads, $\mathcal{H}_{kv}$ the number of kv heads, $\mathcal{D}_{\text{emb}}$ the embedding dimension, $\mathcal{D}_{\text{mlp}}$ the MLP dimension and $\mathcal{S}$ the model size. We fix head dimension at $128$ for all experiments. All models are trained on a 50/50 mix of (i) high-quality web data derived from DCLM (\cite{li2024datacomp}) and (ii) Python code derived from Stackv2 (\cite{lozhkov2024starcoder}).

We apply the telescoping algorithm (Alg.~\ref{alg:tele}) from Sec.~\ref{sec:telescope} to successively refine our hyperparameter sweeps while doubling the model width. At each stage, we reduce the search space geometrically, so that in the final sweep on the $2.40$B model, the mesh points are spaced logarithmically with multiplicative steps of approximately $1.17$ and $1.13$ (comparable to standard mesh sizes, see e.g. the Appendix of \cite{everett2024scaling}). The variation in model performance at the two largest scales is not statistically significant, indicating that the re-meshing could have been stopped earlier without loss of accuracy (i.e. $N_c=16$ in Alg.~\ref{alg:tele}, but could have been chosen smaller, e.g. $N_c=8$). Figure~\ref{fig:telescope-analysis} summarizes the primary results and Figure~\ref{fig:loss_values} in the Appendix depicts the variance in loss per telescoping level, together with a standard power law fit to the minimum loss at each level. For clarity, the variance of the first grid is truncated in the figure. 

We observe that each iteration of the telescoping procedure reduces the variance further; combined with the analysis in Sec.~\ref{sec:telescope}, this supports both precision (from the empirical variance in Eq.~\ref{eq:epsilon}) and accuracy (from theoretical guarantees in Eq.~\ref{eq:opt-shift}). The final full model training on the $3.7$B models achieves a final training loss of $1.61$ nats when trained at the selected weight decay of $\lambda=0.027$ (see Eq.~\ref{eq:muon-step}) and muP base learning rate of $\eta_0 = 0.0612$ (see Sec.~\ref{sec:mup}), outperforming all models at previous widths. We observe excellent agreement with a shifted power law ($R^2\approx 1$), where the exponent for the parameter dependence is $0.31$, close to the Chinchilla exponent of $0.34$ (see Fig.~\ref{fig:loss_values}).

\section{Related Work}
 \label{sec:related}

\paragraph{Dethroning AdamW.}
AdamW \citep{kingma2014adam,loshchilov2017decoupled} has long been the de facto king of the optimizers for large-scale neural network training. Recent research has shown promising results with new optimizers, especially with second-order methods based on non-diagonal preconditioners \citep{gupta2018shampoo,vyas2024soap,jordan2024muon,liu2025muon}, 
but also with first-order methods \citep{chen2023symbolic,shazeer2018adafactor,zhai2022scaling,zhao2025deconstructing}. 
However, these works compare optimizers at fixed compute resources and do not conclusively show that an optimizer is better than AdamW in trading off compute and time resources. We address this limitation by explicitly showing that Muon expands AdamW's compute-time Pareto frontier. 
The notion of the compute-time Pareto frontier is first proposed by \citet{mccandlish2018empirical} for studying batch size dynamics, but to our knowledge has not been used to compare optimizers. 

\paragraph{Impact of the batch size.}
Most research on the impact of batch size on training relies on the idea of critical batch size, broadly construed as the largest batch size that allows for almost linear scaling beyond which we receive diminishing returns \citep{balles2016coupling,goyal2017accurate,mccandlish2018empirical,zhang2019algorithmic,shallue2019measuring,zhang2024does}. While useful, such a point estimate is sensitive to noise and fails to describe the impact of the batch size in the post-critical regime, which can play a critical role in the compute-time tradeoff.  
We present a novel way to continuously measure the batch size advantage of an optimizer in the post-critical regime by monitoring the ratio of token consumptions.

\paragraph{Practical aspects of muP.} 
There are many existing works that consider muP (\cite{ishikawa2023parameterization,blake2024u,haas2024effective,yaida2022meta,dinan2023effective,everett2024scaling,yang2022tensor,dey2024sparse, halverson2024physics, meta2024llama4}), and so it is worth asking in what ways our study differs. Generally speaking, muP has been shown to be theoretically correct, the key insight being that it is possible to relate the infinite-width limit to finite-width network behavior. It has also been empirically demonstrated to work, although existing studies have omitted rigorous investigations and analyses of the errors that we have discussed in this paper. While the primary intent of \cite{everett2024scaling} was to address a particular assumption made in the derivation of muP, another major contribution of the study was the large number of experiments they ran which demonstrate the behavior of muP in practice. \cite{yaida2022meta} discusses different scaling parameterizations and in \cite{dinan2023effective} they discuss their empirical performance. 

\cite{meta2024llama4} highlights a new technique, ``MetaP'', which while not described, emphasizes the need for precise hyperparameter transfer across other scale dimensions (i.e. depth, batch size, sequence length and training steps), which are shown to hold only approximately in \cite{yang2022tensor}. Finally, \cite{ishikawa2023parameterization} proposed techniques for applying muP to second order optimization (i.e. Shampoo and KFAC), however experimental demonstrations of muP together with Muon have yet to have been empirically validated \cite{moonshot2025tweet}. Specifically, despite Muon's close relationship to Shampoo (see. App.~\ref{app:muon-reduction}), we have shown in Sec.~\ref{sec:mup-experiments} that the same muP scaling that is used for AdamW in \cite{yang2022tensor} (see Table~\ref{table:params}) works for Muon. Due to the considerable similarities between Muon's hyperparameter transfer properties and muP's hyperparameter transfer properties discussed in \cite{bernstein2025deriving} our observation of Muon's simple compatibility with muP is consistent with existing literature.

\section{Conclusion}
\label{sec:conclusion}

This paper has addressed two practical questions that arise in language model pretraining: (1) which optimizer delivers the best tradeoff between compute and time resources, and (2) how to tune that optimizer without expending prohibitive compute.  
For the first question, we have shown that Muon expands AdamW's Pareto frontier on the compute-time plane, enlarging the practitioner's flexibility in resource allocation. We have given an account for this advantage by analyzing the token ratio \eqref{eq:token-ratio}, which measures the relative data efficiency of Muon at large batch sizes. Concretely, Muon requires $10$–$15\,\%$ fewer tokens than AdamW to reach an identical loss and converts these savings into faster wall-clock convergence, with the advantage staying constant or growing as the batch size increases. The findings are validated by extensive experiments across five model sizes ($100$M–$4$B parameters), two data modalities, and an order of several decades of variation in global batch size. These results establish Muon as a drop-in successor to AdamW for second-order optimization at scale.

For the second question, we have given the first affirmative answer to the open question posed by \citet{moonshot2025tweet} and \citet{liu2025muon}: does the maximal-update parametrization (muP) remain valid when used together with Muon?  Our experiments confirm that muP transfers hyperparameters cleanly up to $3.7$B-parameter models at sequence length $8192$, for both learning rate and coupled weight decay.  Building on this observation, we have introduced a telescoping hyperparameter sweep that contracts the search grid at each width doubling, bounding tuning overhead by a factor of $O(C\log N)$ where $N$ is the final model width and $C$ is the cost of training the full model, while quantifying all known sources of transfer error (Eq.~\ref{eq:opt-shift}, Eq.~\ref{eq:epsilon}, Fig.~\ref{fig:telescope-analysis}).  In practice, more than $20$\% of the total compute budget is now devoted to the final, full-scale training run while guaranteeing near-optimal hyperparameters.

Taken together, these contributions give a unified recipe: Muon optimization, muP scaling, and telescoping hyperparameter transfer. The recipe delivers strictly superior data efficiency, shorter training time, and near-negligible tuning cost compared with the AdamW baseline. The evidence presented here promotes Muon from a promising alternative to a robust, drop-in replacement for AdamW, and converts second-order optimization---long regarded as computationally extravagant---into a practical default for industry-scale language model pretraining. 

\bibliographystyle{natbib}
\bibliography{mybib}

\appendix
\newpage
\section{Contributions}

\textbf{Core Contributors} 

\vspace{2mm}
Ishaan Shah$^\star$\\
Anthony M. Polloreno$^\star$\\ 
Karl Stratos$^\star$\\ 
Philip Monk\\
Ashish Vaswani

\vspace{8mm}
\textbf{Contributors} 

\vspace{2mm}
Adarsh Chaluvaraju\\
Andrew Hojel\\
Andrew Ma\\
Anil Thomas\\
Ashish Tanwer\\
Darsh J Shah\\
Khoi Nguyen\\
Kurt Smith\\
Michael Callahan\\
Michael Pust\\
Mohit Parmar\\
Peter Rushton\\
Platon Mazarakis\\
Ritvik Kapila\\
Saurabh Srivastava\\
Somanshu Singla\\
Tim Romanski\\
Yash Vanjani

\def\thefootnote{$\star$}\footnotetext{Equal contribution}
\newpage

\section{Reduction of Shampoo and Soap to Muon}
\label{app:muon-reduction}

It is well known that Shampoo \citep{gupta2018shampoo} without momentum is equivalent to Muon \citep{jordan2024muon}. 
At its core, Shampoo uses the following gradient transformation 
\begin{align*}
    O_t &= \expect{G G^\top }^{-1/4} G_t \expect{G^\top G}^{-1/4}
\end{align*}
where $G_t \in \R^{m \by n}$ is the gradient on the current batch and $G \in \R^{m \by n}$ is the gradient on a random batch. The second moments (assumed to be full rank) can be estimated in an online fashion using bias-corrected exponential moving average (EMA). If we instead consider a point estimate $\expect{G G^\top } \approx G_t G_t^\top$ and $\expect{G^\top G} \approx G_t^\top G_t$, we have 
\begin{align*}
    O_t &\approx (G_t G_t^\top)^{-1/4} G_t (G_t^\top G_t)^{-1/4} = U\Sigma^{-1/2} U^\top U \Sigma V^\top V \Sigma^{-1/2} V^\top = U V^\top
\end{align*}
where $G_t = U \Sigma V^\top$ denotes an SVD. 
A less well-known connection is between Soap \citep{vyas2024soap} and Muon. Soap performs Adam in a second-order basis. Specifically, let $\expect{G G^\top } = Q_L \Lambda_L Q_L^\top$ and $\expect{G G^\top } = Q_R \Lambda_R Q_R^\top$ denote the eigendecomposition of the second moments. Soap uses the following gradient transformation
\begin{align*}
    \wb{G}_t &= Q_L^\top G_t  Q_R \\
    M_t &= \mathrm{UpdateEMA}\paren{\wb{G}_t} \\
    V_t &= \mathrm{UpdateEMA}\paren{\wb{G}_t^{\langle 2 \rangle}} \\
    \wb{O}_t &= \frac{M_t}{\sqrt{V_t}} \\
    O_t &= Q_L \wb{O}_t Q_R^\top
\end{align*}
where $A^{\langle 2 \rangle}$ is the elementwise square of $A$. Consider again computing $Q_L, Q_R$ from a point estimate. Then $Q_L = U$ and $Q_R = V$ since $G_t G_t^\top = U \Sigma^2 U^\top$ and $G_t^\top G_t = V \Sigma^2 V^\top$. Further computing $M_t, V_t$ from a point estimate, we have (assuming $m \leq n$)
\begin{align*}
    \wb{G}_t &= \Sigma \\
    M_t &= \Sigma \\
    V_t &= \Sigma^2 \\
    \wb{O}_t &= I_{m \by m} \\
    O_t &= U V^\top
\end{align*}
Thus Shampoo and Soap can be reduced to Muon under these simplifying assumptions.
In particular, the motivation for Shampoo as an implicit approximation of the inverse square root of the empirical Fisher matrix (which itself can be viewed as an approximation of the inverse Hessian) transfers to Muon.

\vspace{-0mm}
\begin{figure}[t!]
\begin{center}
\begin{tabular}{cc}
\hspace{-1mm}\includegraphics[width=6.9cm]{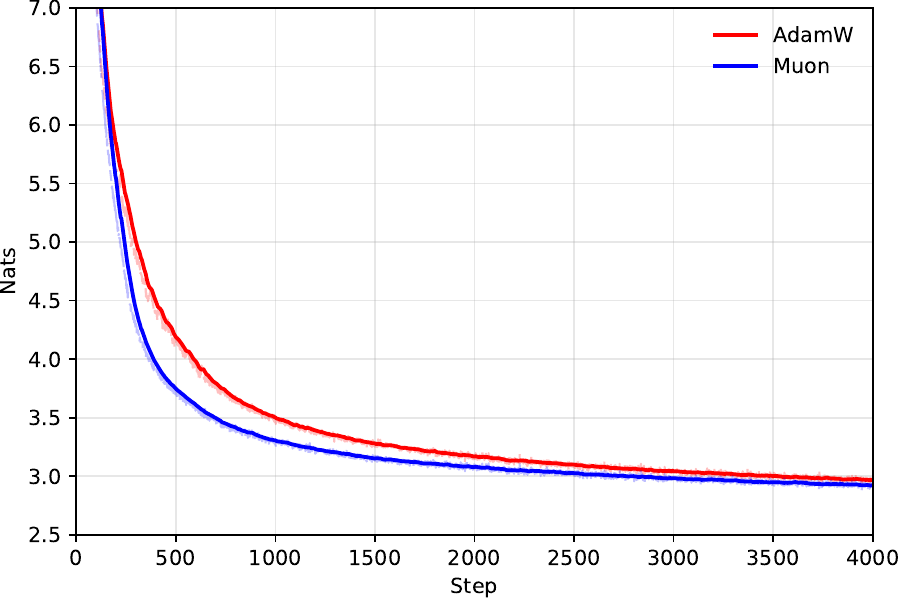}
& 
\hspace{-4mm}\includegraphics[width=6.9cm]{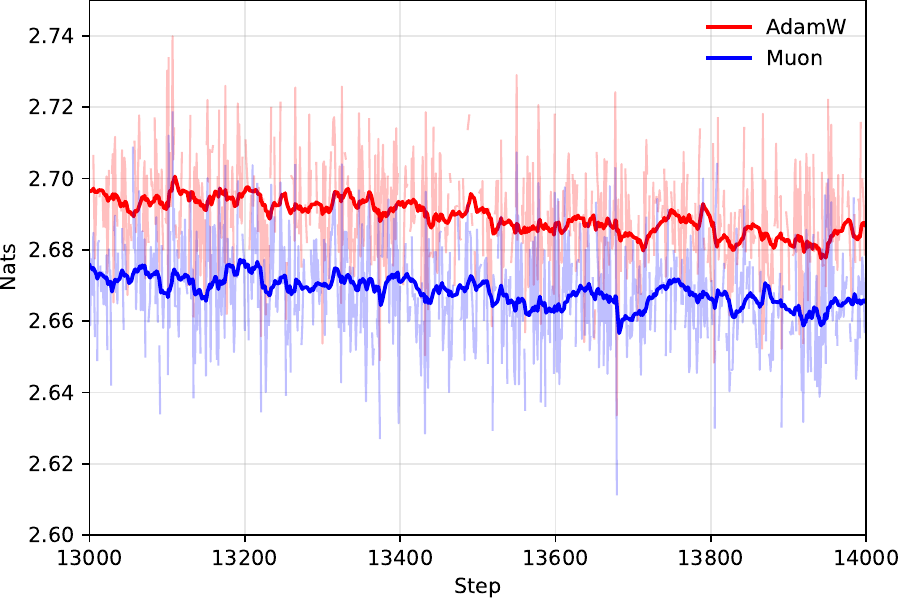} \\
\hspace{-1mm}\includegraphics[width=6.9cm]{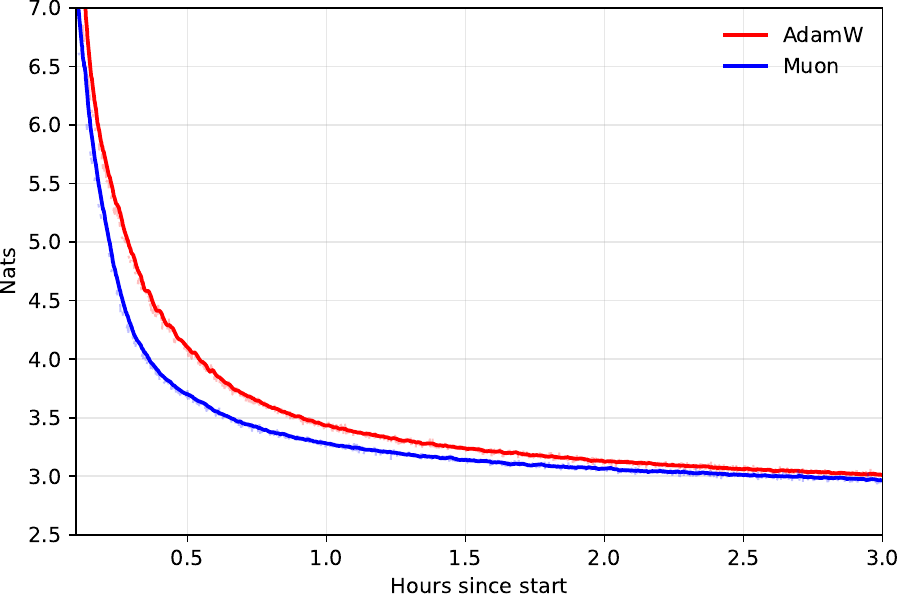}
& 
\hspace{-4mm}\includegraphics[width=6.9cm]{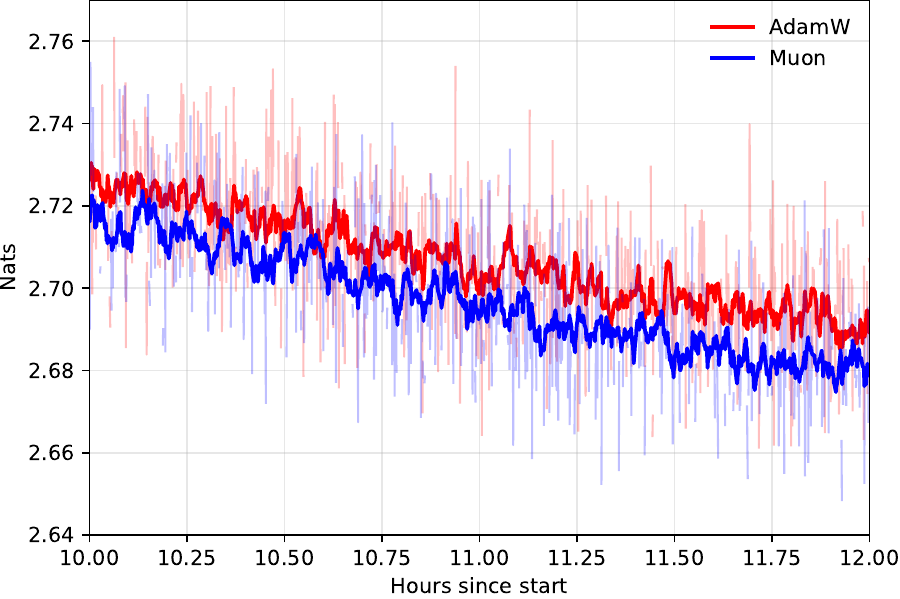} \\
\end{tabular}
\end{center}
\caption{Comparison of the best 1B model training runs on DCLM under AdamW and Muon in steps and wall time. (Top) Muon has a consistently lower training loss compared to Adam given the same number of steps. The difference does not vanish at the end of training ($0.02$ nats at $2.5\times$ Chinchilla-optimal). (Bottom) Muon obtains a target loss faster than Adam.}
\label{fig:prelim}
\end{figure}

\section{Initial Verification of Muon's Performance}
\label{app:initial-verification}

We find that Muon consistently achieves a lower training loss and faster convergence across different model sizes and datasets. We give a concrete example with 1B model size on DCLM in Figure~\ref{fig:prelim}. We take the best training runs (i.e., the hyperparameter configurations resulting in the lowest final loss) under Adam and Muon and plot the training loss as a function of steps and hours (smoothed using EMA with coefficient 0.95).
It is clear that Muon strictly lower bounds AdamW on these curves, even to the end of training far beyond the Chinchilla-optimal budget (i.e., there is no crossover). 
It is notable that the wall time advantage holds with our minimal Muon implementation and may further improve with a more sophisticated version (e.g., using lower precision for the first moment).
Our finding strengthens and expands the existing evidence on Muon's promising utility for pretraining \citep{jordan2024muon,liu2025muon}.

\section{Deeper Analysis of the Critical Batch Size}
\label{app:critical-ours}

Increasing the batch size is one of the easiest ways to accelerate training on data parallel hardware.
A great deal of research focuses on the idea of ``critical'' batch size, which is broadly defined as the largest batch size that allows for almost linear or ``perfect'' scaling (e.g., doubling the batch size halves the number of steps) beyond which we receive diminishing returns.

Despite the simple intuition, formalizing the critical batch size can be nontrivial \citep{mccandlish2018empirical}. 
A more direct approach is to measure the number of steps $S_L(B)$ required to reach a target loss $L$ as a function of the batch size $B$. 
Let $T_L(B) = B \times S_L(B)$ denote the corresponding number of tokens consumed. We can define the critical batch size as a value $B_L^\star$ such that (1) $T_L(B) =  T_L^\star$ remains constant for $B \leq B_L^\star$ where $T_L^\star$ represents the minimum number of tokens required to reach loss $L$, and (2) $T_L(B) = \mathrm{inc}_L(B)$ is some increasing function for $B > B_L^\star$.\footnote{The specific form of $\mathrm{inc}_L(B)$ in the post-critical regime is not prescribed (other than generally increasing). Appendix~\ref{app:critical-toy} gives a simple parametric model for completeness.} In the perfect scaling regime, we have $S_L(B) = T_L^\star B^{-1}$ which is linearly decreasing with slope $-1$ in log-log scale. 
Researchers often gauge this ``phase transition'' (i.e., kink) from an empirical plot, and argue that an optimizer enjoys a larger critical batch size if the transition seems to happen at a larger batch size \citep{zhang2019algorithmic,vyas2024soap}.

\vspace{-0mm}
\begin{figure}[t!]
\begin{center}
\begin{tabular}{ccc}
\hspace{-1mm}\includegraphics[width=4.5cm]{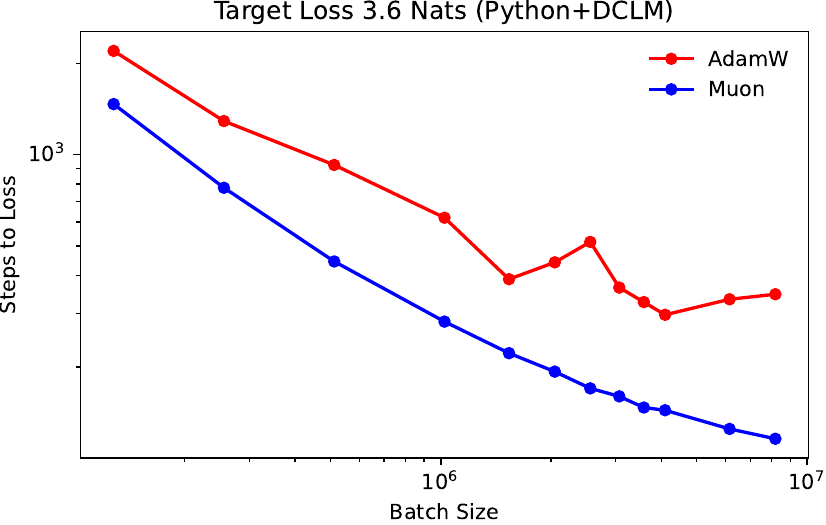}
& 
\hspace{-4mm}\includegraphics[width=4.5cm]{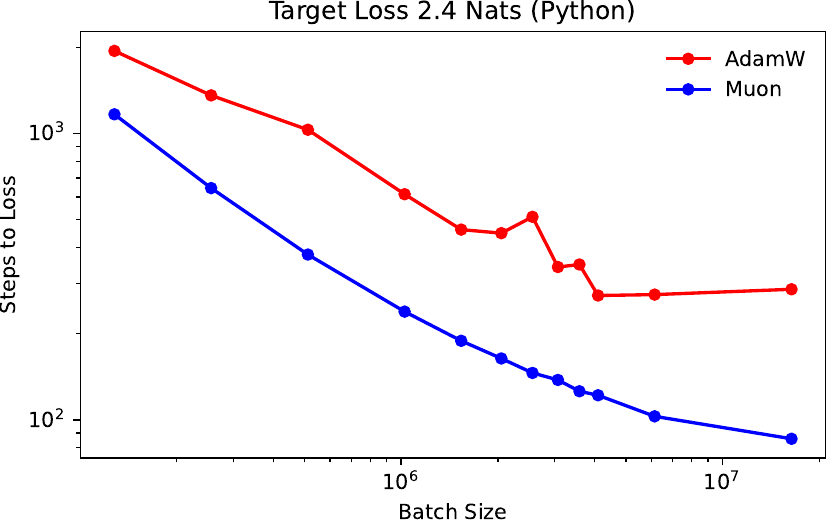}
&
\hspace{-4mm}\includegraphics[width=4.5cm]{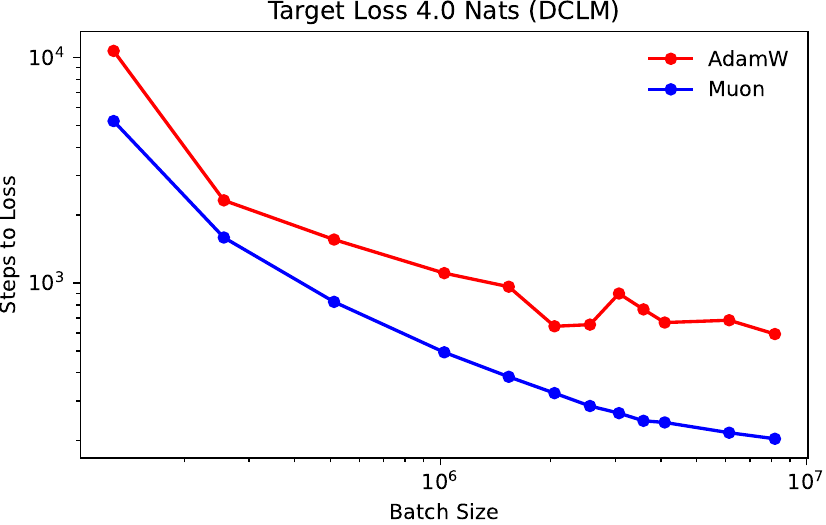}
\\
\hspace{-1mm}\includegraphics[width=4.5cm]{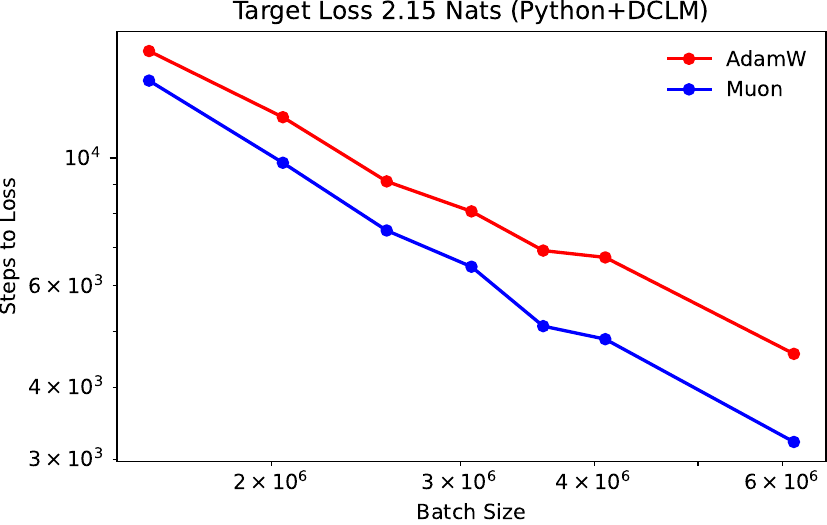}
& 
\hspace{-4mm}\includegraphics[width=4.5cm]{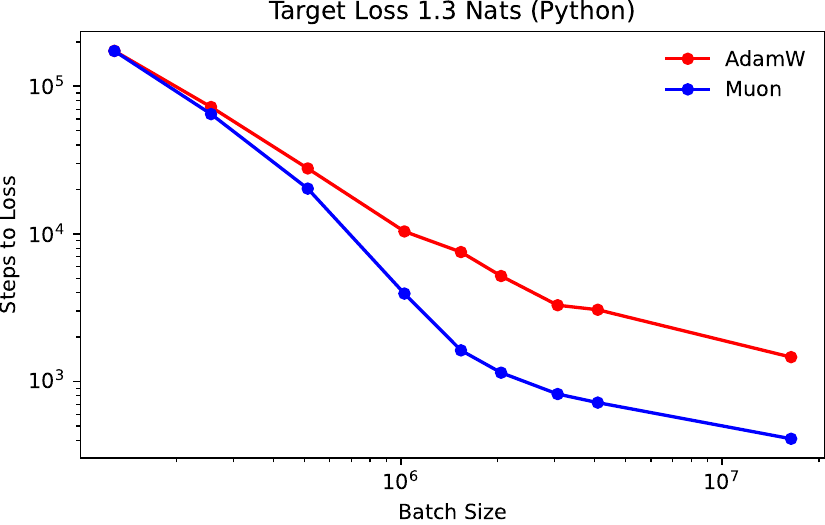} 
& 
\hspace{-4mm}\includegraphics[width=4.5cm]{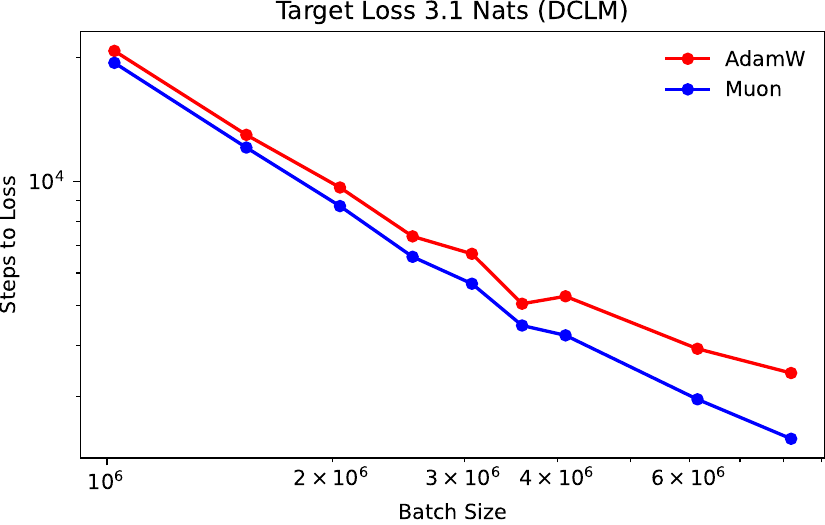}
\end{tabular}
\end{center}
\caption{Steps-to-loss vs batch size plots with 500M parameter models for target losses near convergence (log-log scale).}
\label{fig:steps-batch}
\end{figure}
Figure~\ref{fig:steps-batch} shows the step curves for AdamW and Muon (in log-log scale).
While the Muon curve lower bounds the AdamW curve in general, they are limiting for the purpose of critical batch size study for the following reasons. First, they do not always exhibit a clear phase transition (e.g., there may be no major kink, or there may be multiple), particularly at a low target loss. 
Second, the semantics of the plot is tightly coupled with the slope. If the slope is larger than $-1$, it implies that we should use a smaller batch size regardless of any kink since we are already in the regime of diminishing returns. Conversely, if the slope is smaller than $-1$, there can be several kinks but none of them is a critical batch size. Thus without reporting the actual slope value, the visual presence of phrase transitions alone can be misleading.

\subsection{Token-optimal batch size}
\label{sec:token-efficiency}

To facilitate a more effective measurement of optimal batch size, we can examine the tokens-to-loss $T_L(B)$ rather than steps-to-loss $S_L(B)$.
By definition, the tokens-to-loss curve is flat in the perfect scaling regime and increasing in the post-critical regime (Figure~\ref{fig:critical-figures} left).  We may then consider a ``token-optimal'' batch size as
\begin{align}
    B_{\mathrm{tok},L} = \max \myset{B:\; T_L(B) = \min_{B'} T_L(B')} \label{eq:tok-opt}
\end{align}
(i.e., the largest data-optimal batch size).
Unlike the kinks in the steps-to-loss plot, $B_{\mathrm{tok},L}$ is always uniquely defined and allows for convenient measurement. 
In an idealized setting where the critical batch size assumption holds, $B_{\mathrm{tok},L}$ coincides with  $B_L^\star$.

\begin{figure}[t!]
\begin{center}
\begin{tabular}{cc}
\hspace{-1mm}\includegraphics[width=6.9cm]{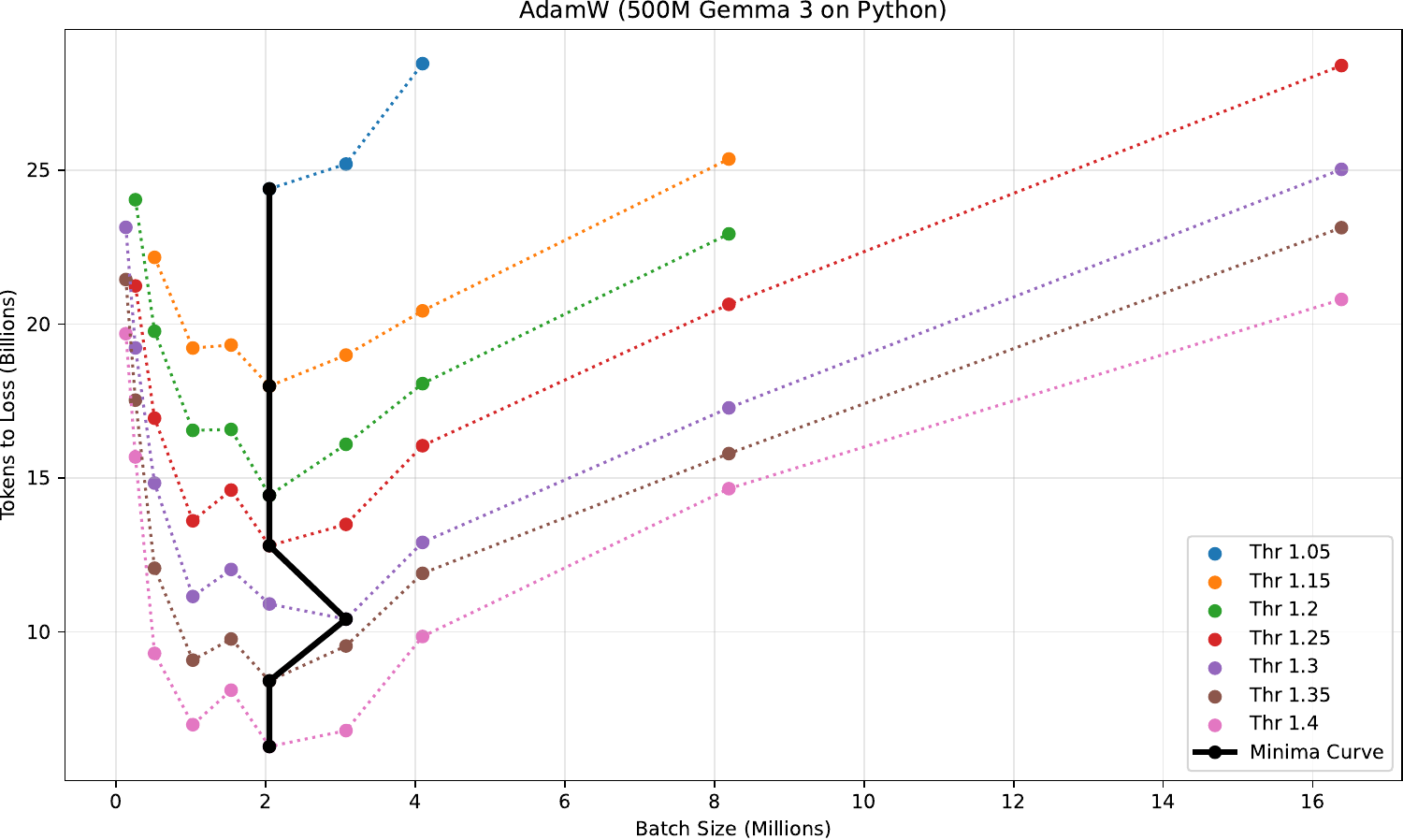}
& 
\hspace{-4mm}\includegraphics[width=6.9cm]{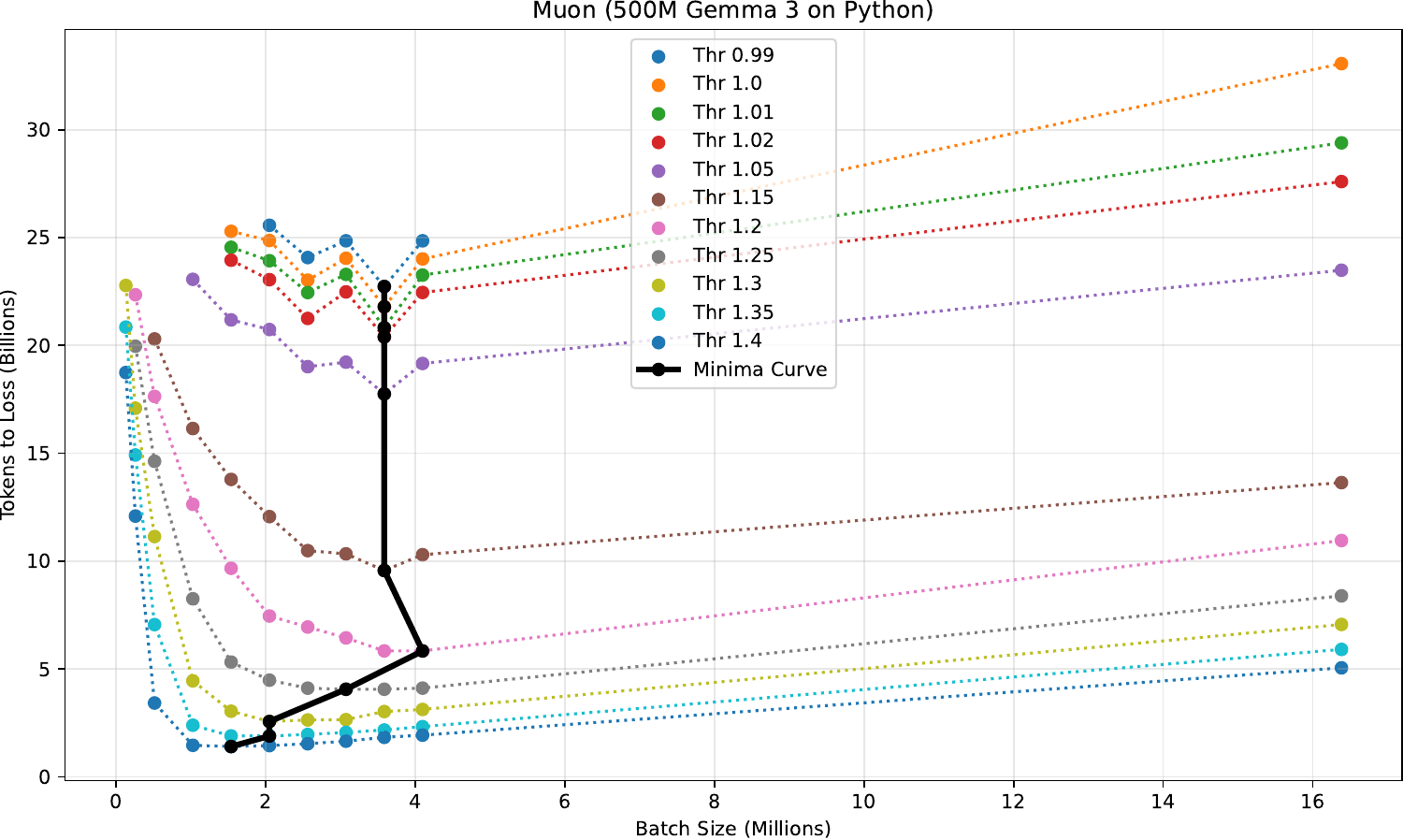}\\
\hspace{-1mm}\includegraphics[width=6.9cm]{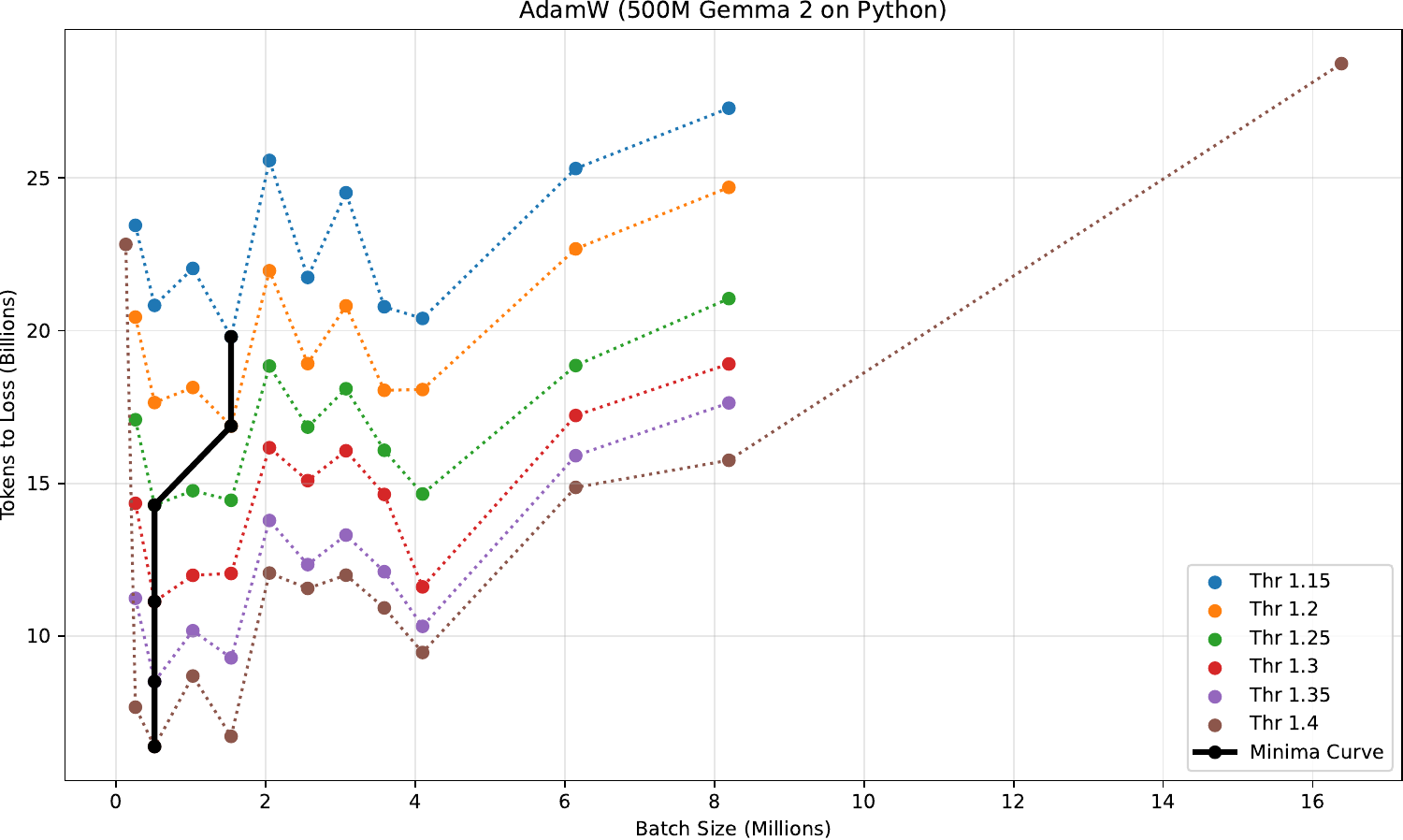}
& 
\hspace{-4mm}\includegraphics[width=6.9cm]{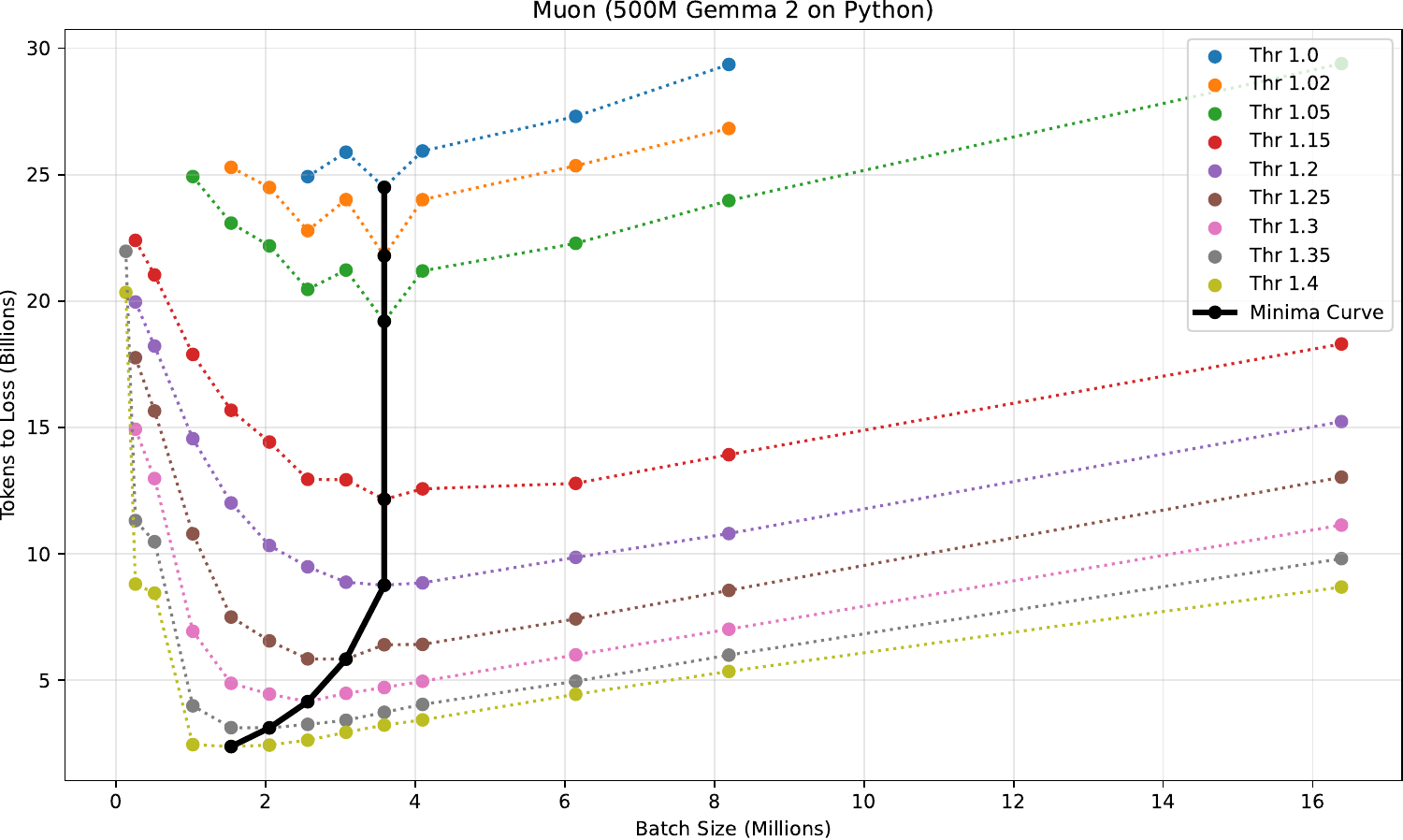}
\end{tabular}
\end{center}
\caption{Token-optimal batch size plots: architecture ablation (Gemma 3 vs Gemma 2)}
\label{fig:token-eff-plots-arch}
\end{figure}

\begin{figure}[t!]
\begin{center}
\begin{tabular}{cc}
\hspace{-1mm}\includegraphics[width=6.9cm]{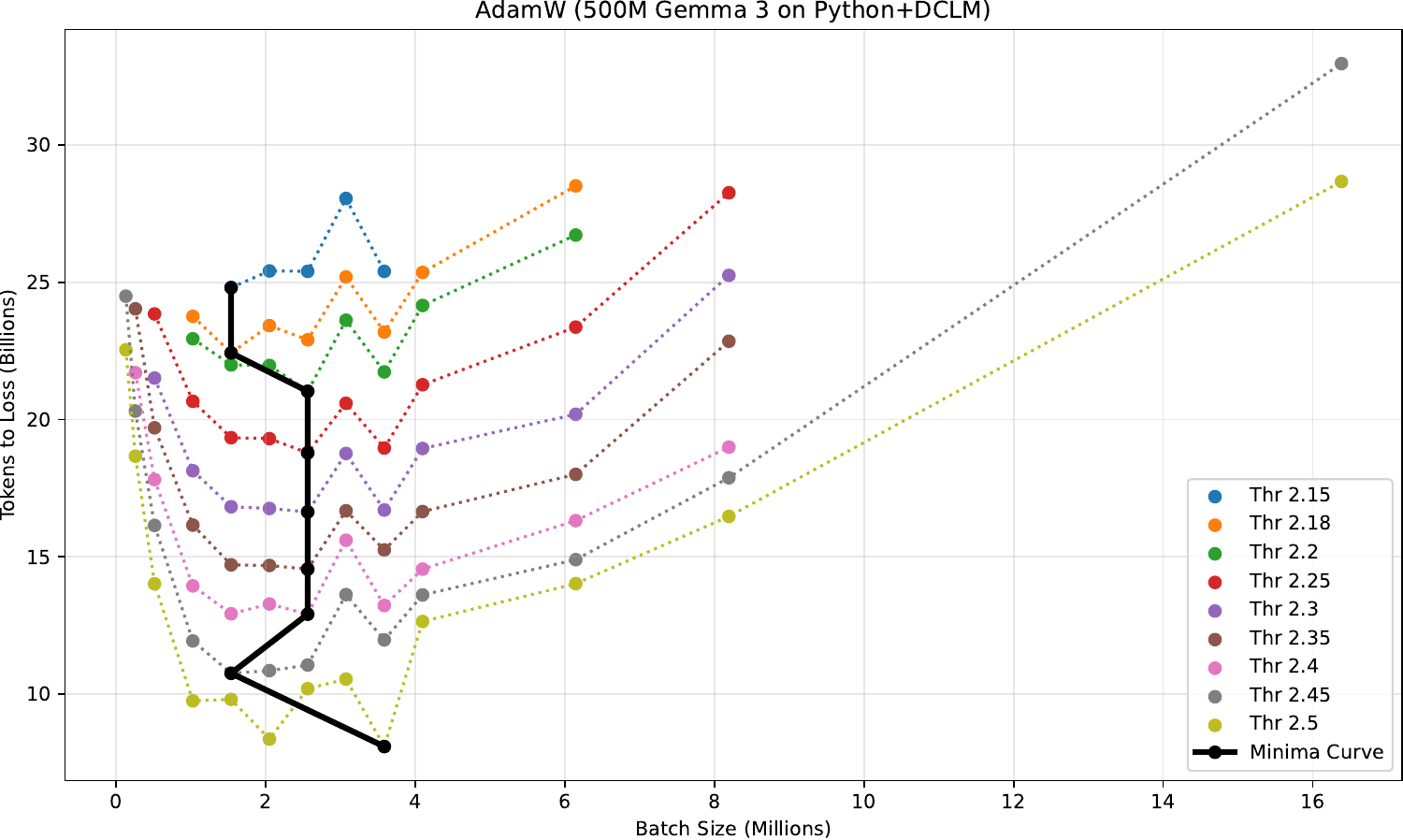}
& 
\hspace{-4mm}\includegraphics[width=6.9cm]{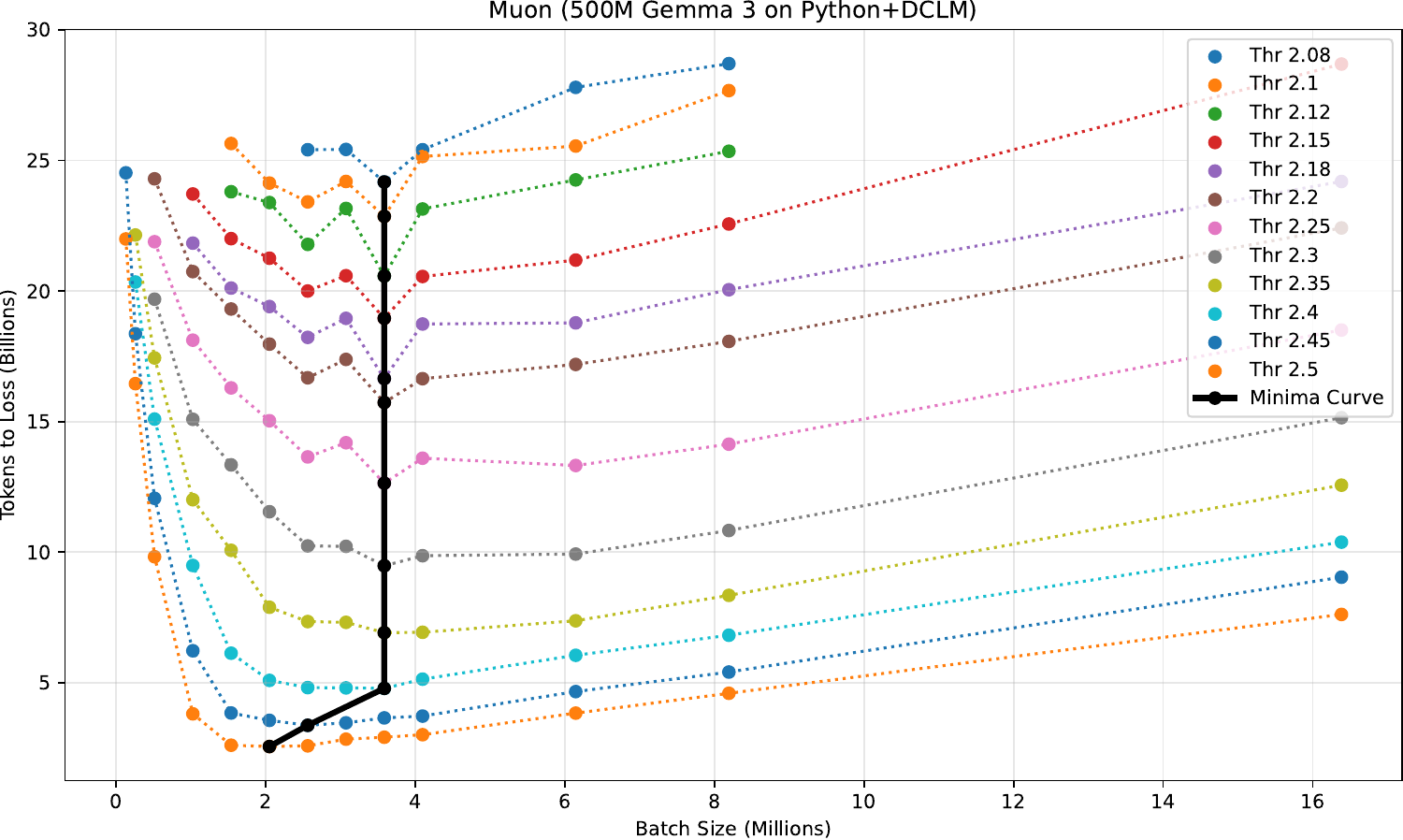} \\
\hspace{-1mm}\includegraphics[width=6.9cm]{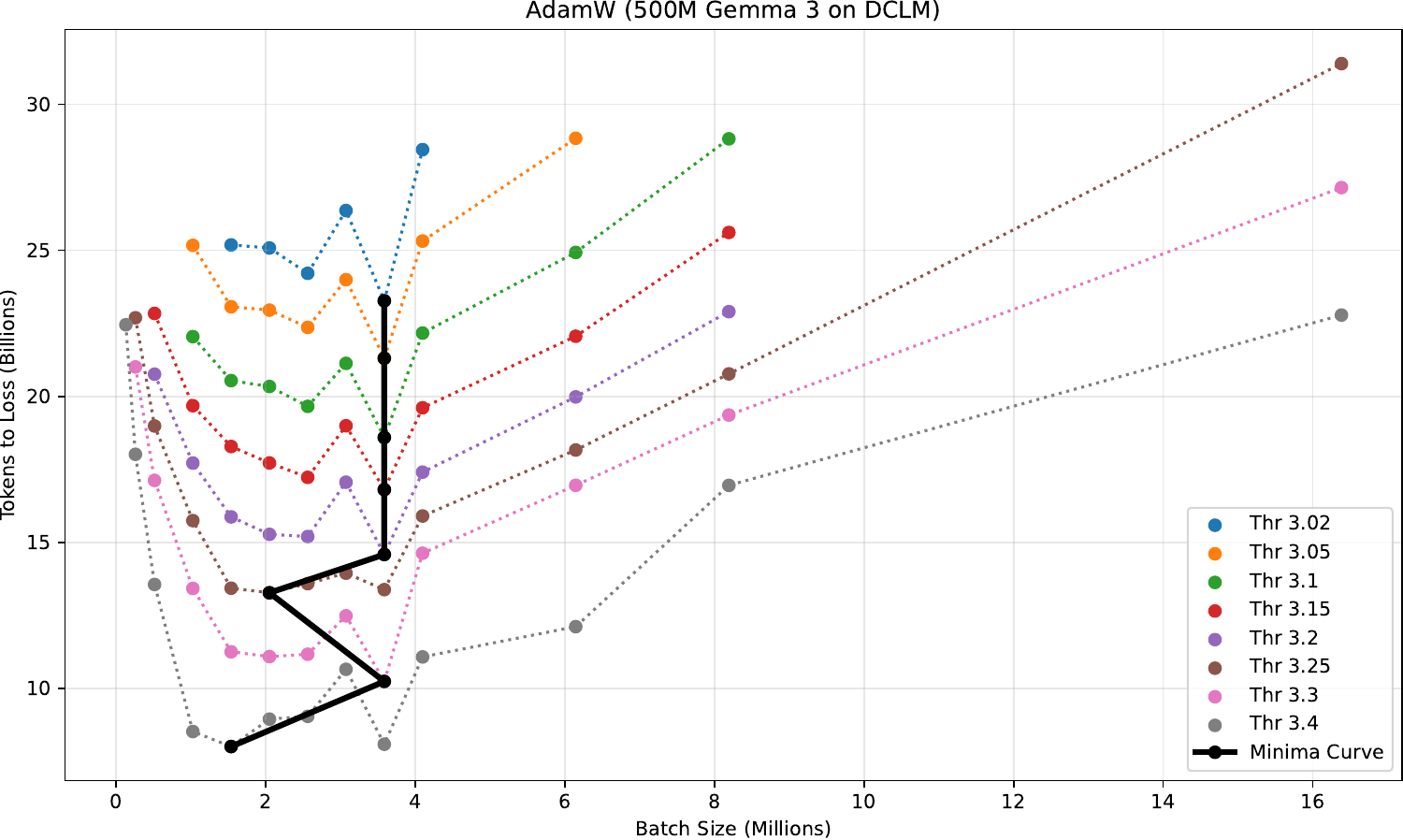}
& 
\hspace{-4mm}\includegraphics[width=6.9cm]{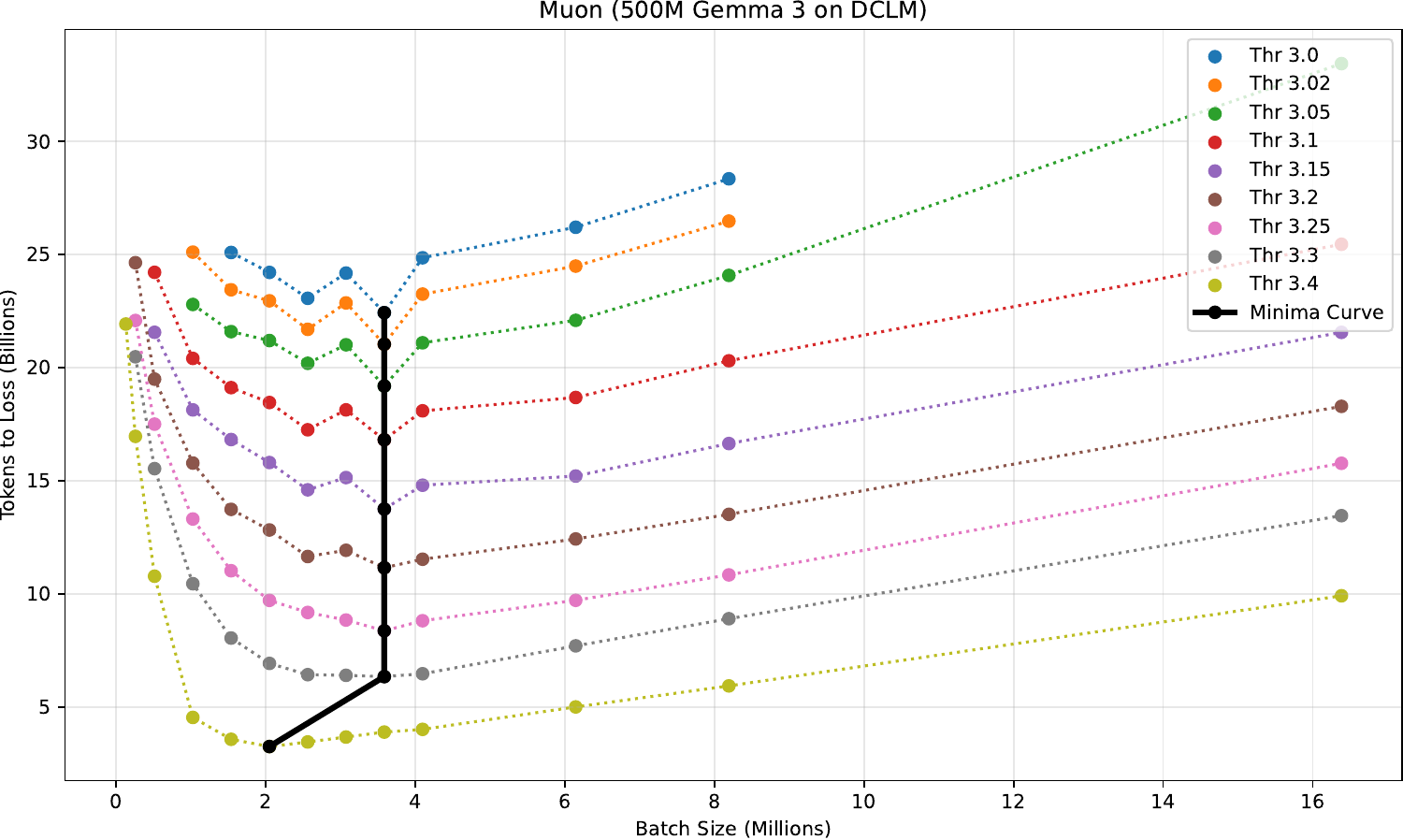}
\end{tabular}
\end{center}
\caption{Token-optimal batch size plots: dataset ablation (Python+DCLM and DCLM)}
\label{fig:token-eff-plots-data}
\end{figure}

\begin{figure}[t!]
\begin{center}
\begin{tabular}{cc}
\hspace{-1mm}\includegraphics[width=6.9cm]{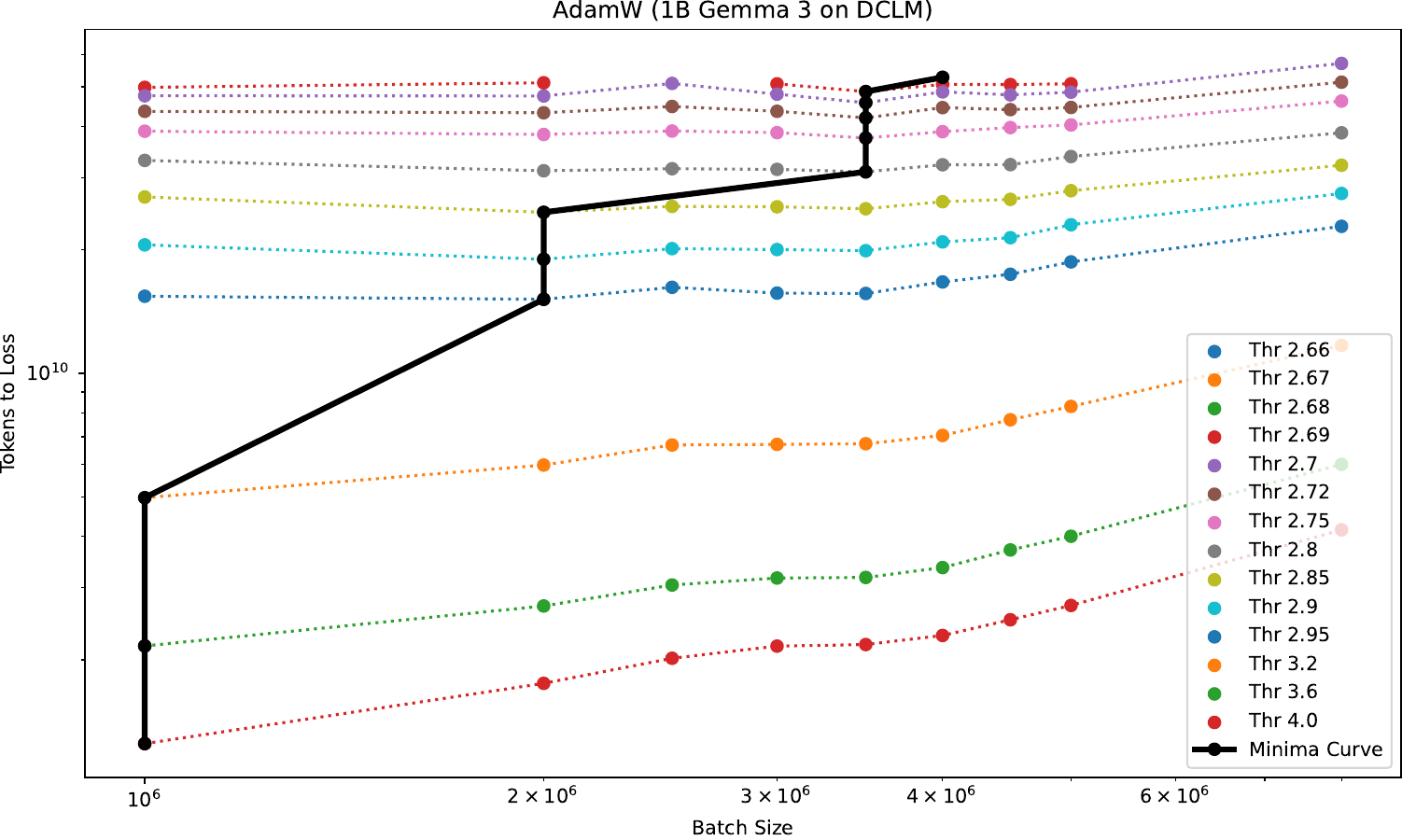}
& 
\hspace{-4mm}\includegraphics[width=6.9cm]{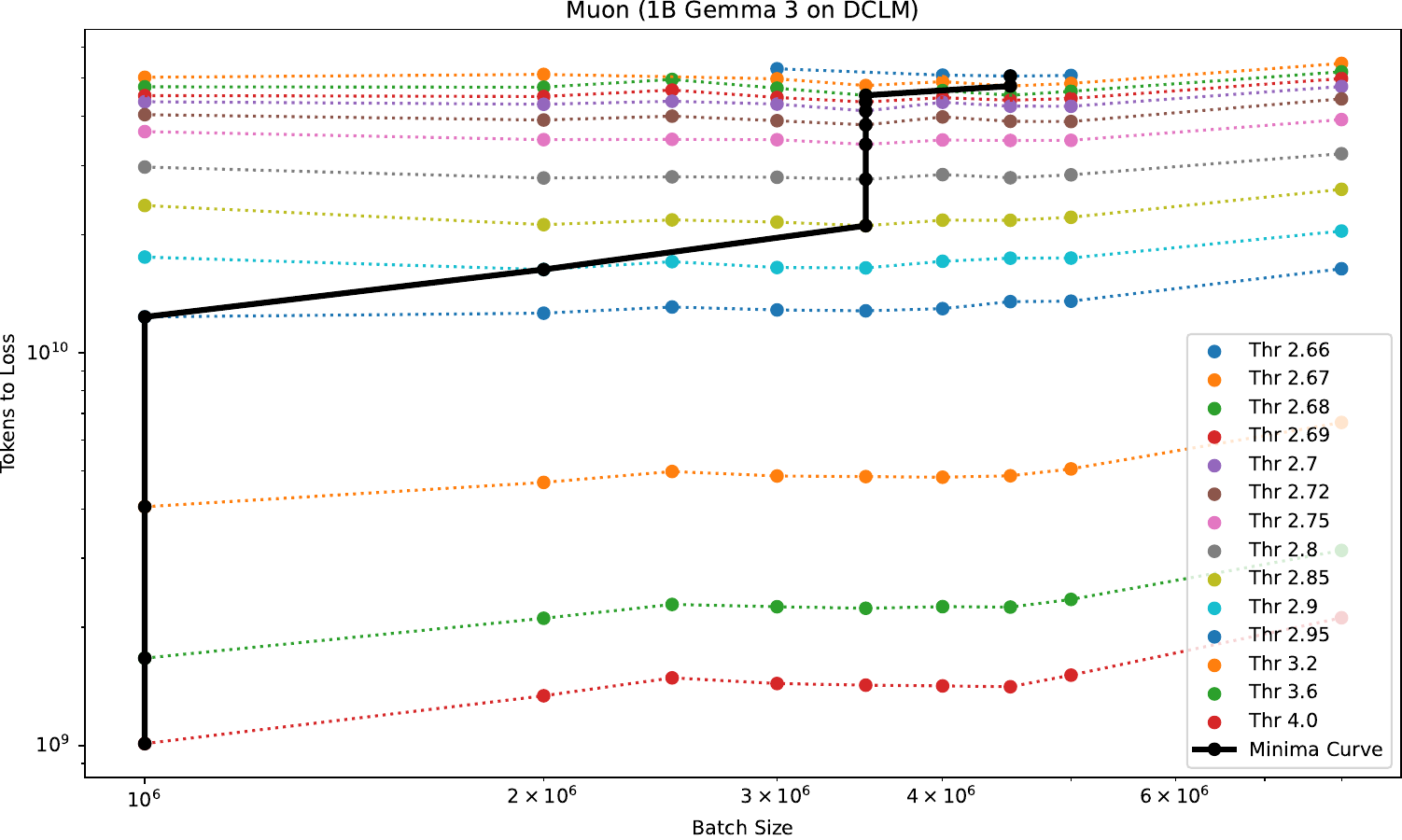} \\
\hspace{-1mm}\includegraphics[width=6.9cm]{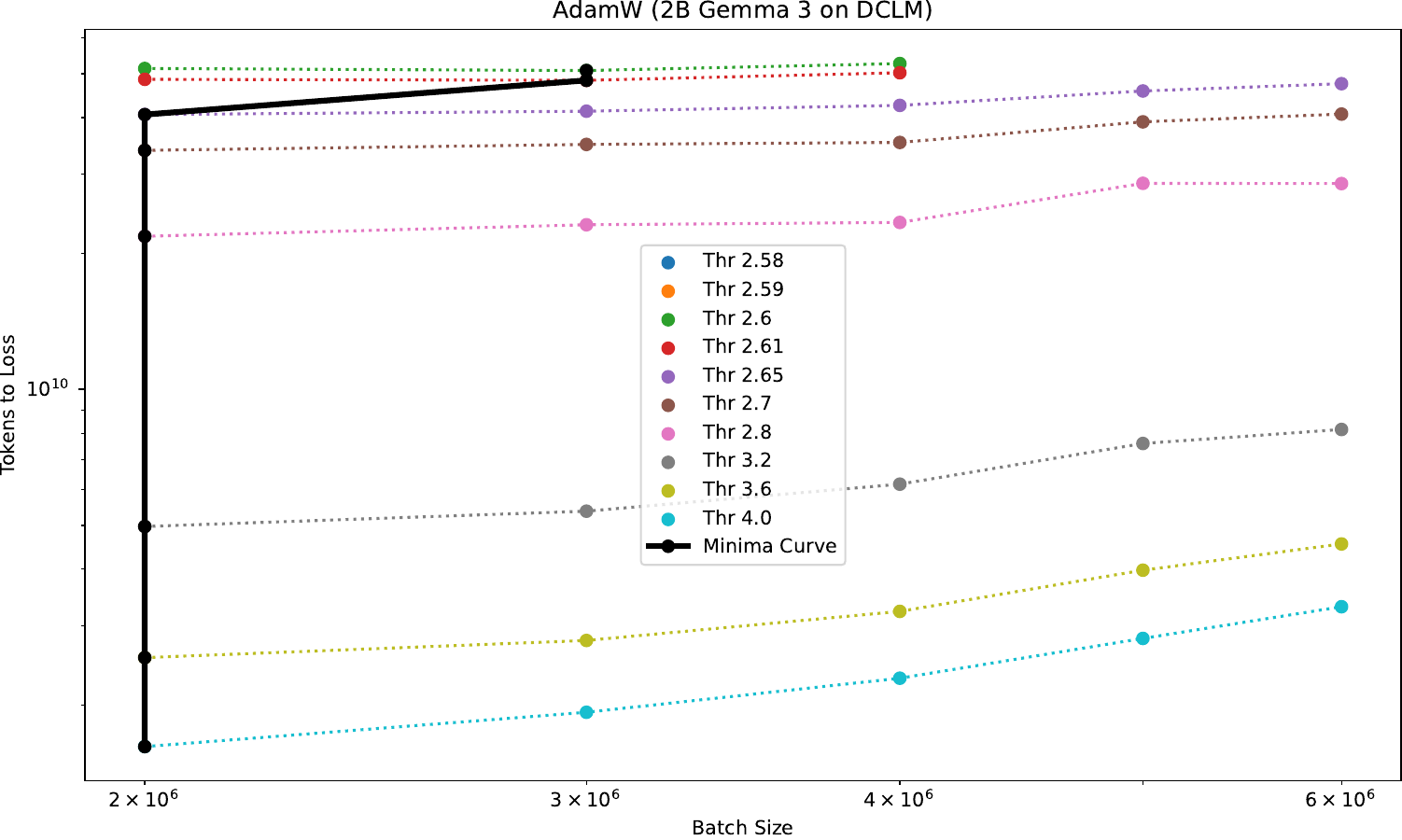}
& 
\hspace{-4mm}\includegraphics[width=6.9cm]{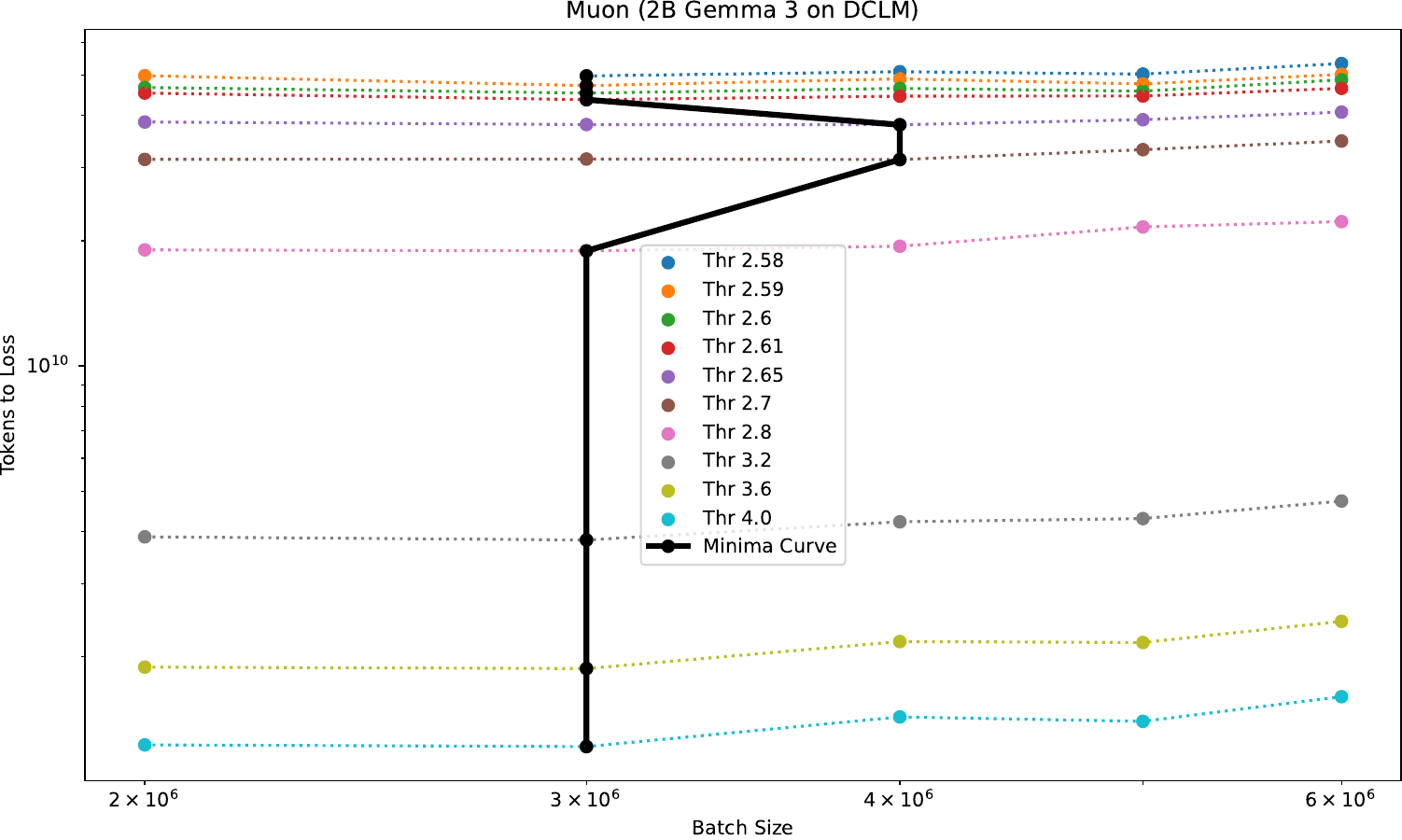} \\
\hspace{-1mm}\includegraphics[width=6.9cm]{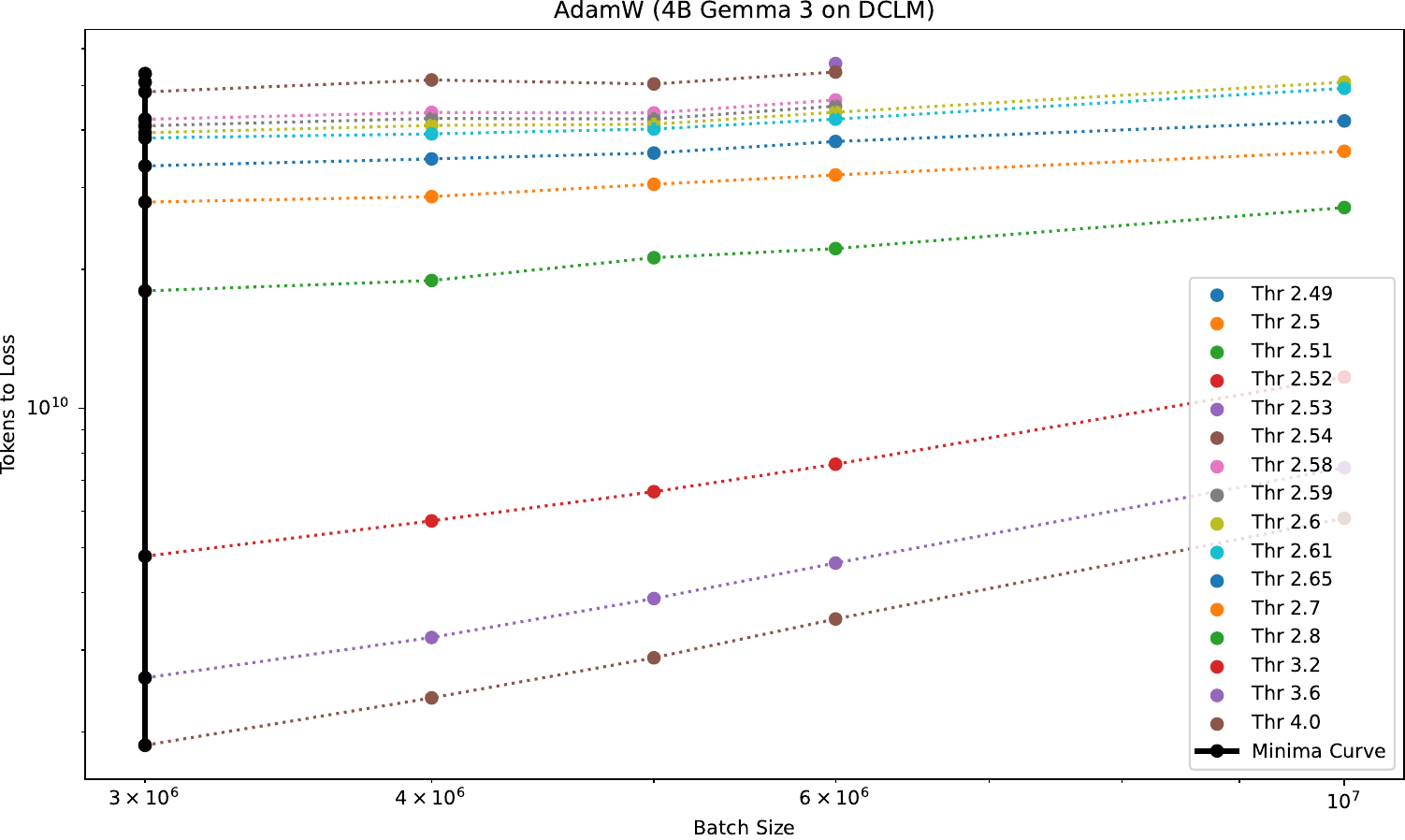}
& 
\hspace{-4mm}\includegraphics[width=6.9cm]{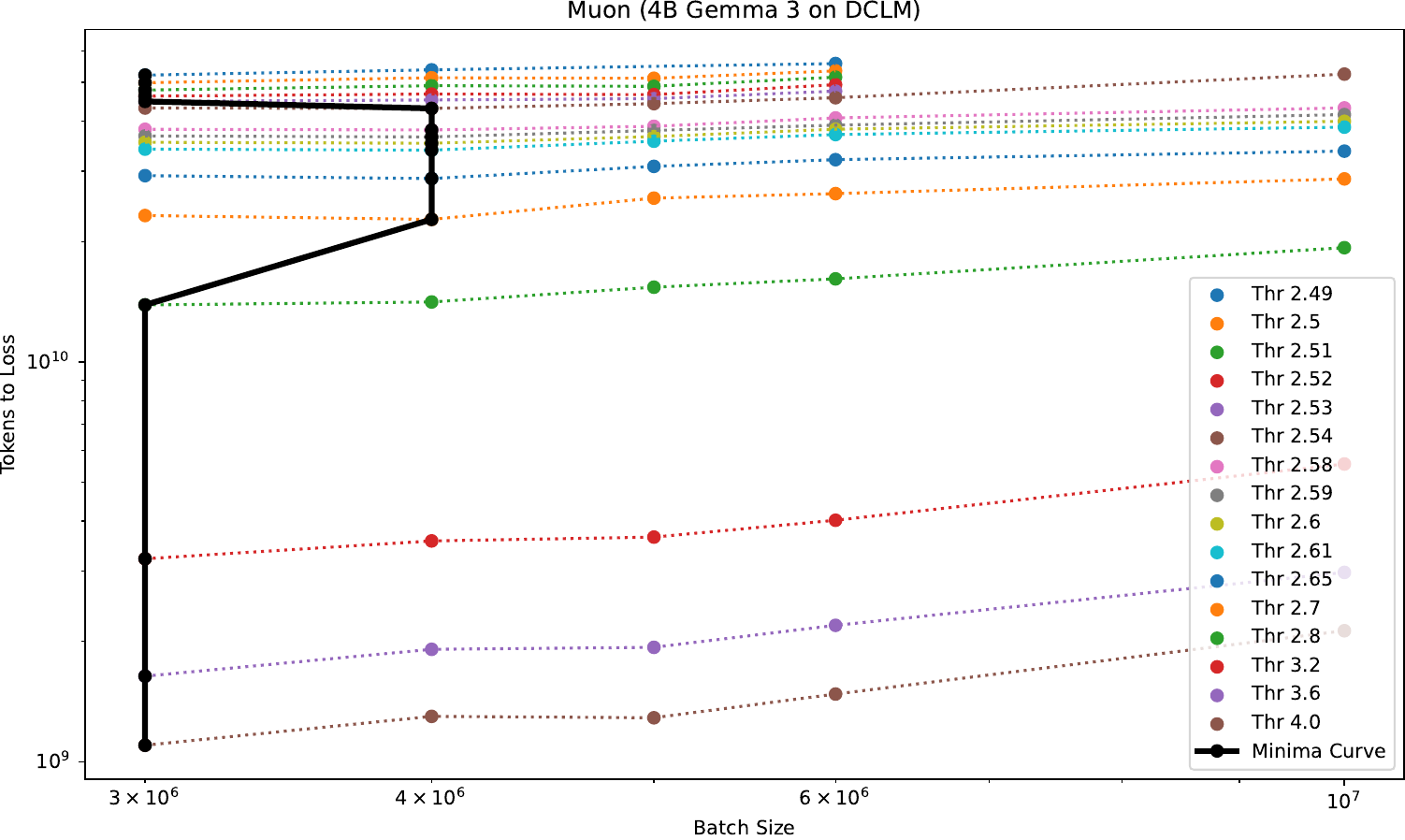} 
\end{tabular}
\end{center}
\caption{Token-optimal batch size plots: model size ablation (1B, 2B, and 4B)}
\label{fig:token-eff-plots-model-size}
\end{figure}

We can monitor the evolution of \eqref{eq:tok-opt} throughout training by plotting it across various loss thresholds near convergence.
Figure~\ref{fig:token-eff-plots-arch}, \ref{fig:token-eff-plots-data}, and \ref{fig:token-eff-plots-model-size}  shows such plots for AdamW and Muon ablating the architecture, dataset, and model size. We see that the token-optimal batch size increases during training for both optimizers, which is consistent with the predicted behavior of critical batch size \citep{mccandlish2018empirical}. 
Muon generally achieves a larger token-optimal batch sizes across these ablations.

\vspace{-5mm}

While useful, the token-optimal batch size is still insufficient for our purpose of fully characterizing the batch size behavior of an optimizer.
First, it is brittle under noise. We can easily construct two nearly identical tokens-to-loss curves where one has a much larger token-optimal batch size, but the advantage flips with an infinitesimal perturbation. Second, much more seriously,  
it does not capture the batch size benefit in the over-large batch size regime.
For instance, on DCLM (Figure~\ref{fig:token-eff-plots-data} bottom row), both AdamW and Muon have the same token-optimal batch size of 3.5M at loss 3.25. However, the token consumption increases much more slowly for Muon compared to AdamW as we further push the batch size beyond 3.5M. 
We capture this large-batch data efficiency behavior with the token ratio \eqref{eq:token-ratio}.

\begin{figure}[t!]
\begin{center}
\begin{tabular}{cc}
\hspace{-3mm}\includegraphics[width=6cm]{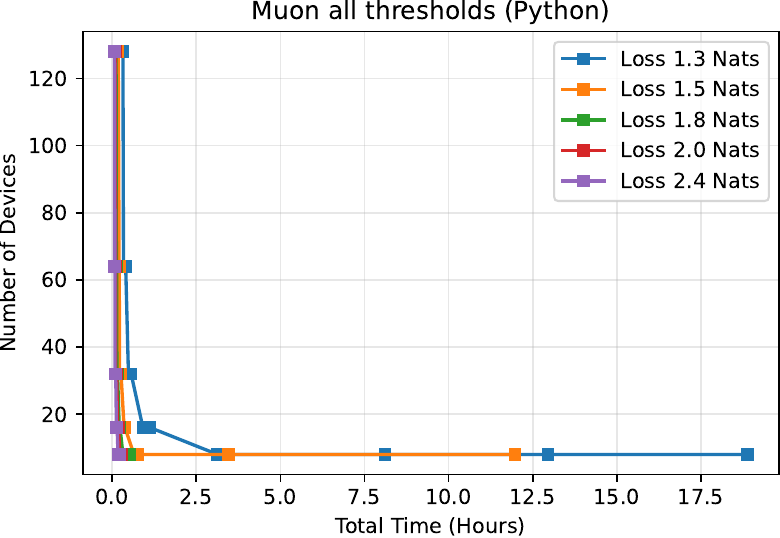}
& 
\hspace{-4mm}\includegraphics[width=6cm]{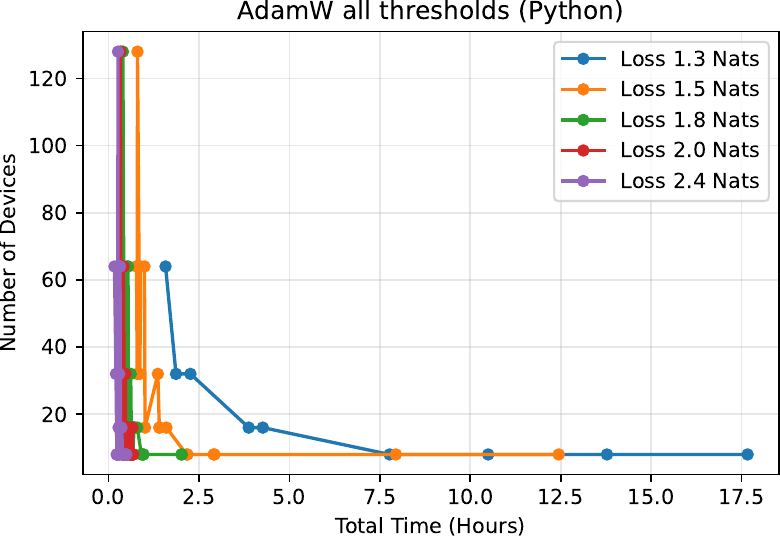} \\
\hspace{-3mm}\includegraphics[width=6cm]{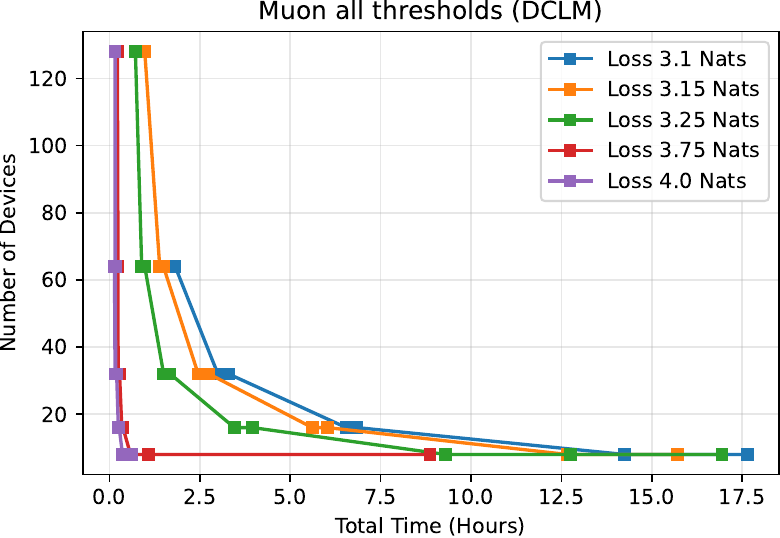}
& 
\hspace{-4mm}\includegraphics[width=6cm]{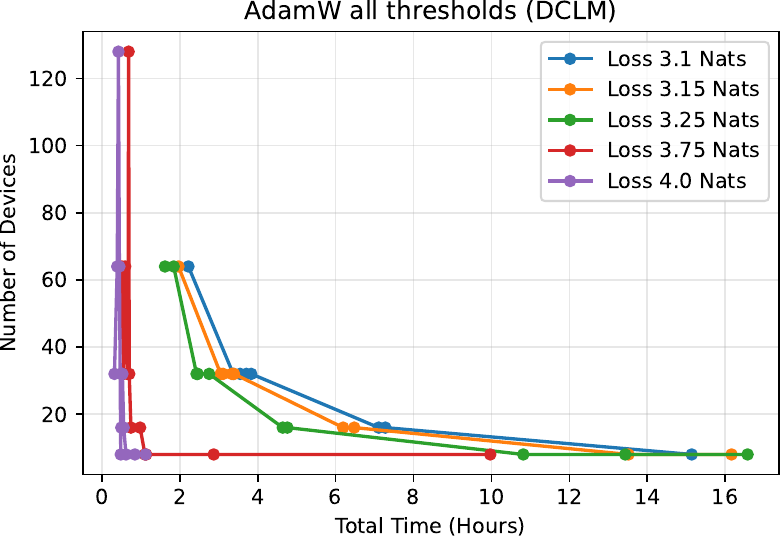}
\end{tabular} 
\end{center}
\caption{Compute-time tradeoff plots for all thresholds between Muon vs AdamW: Python (top) and DCLM (bottom).}
\label{fig:tradeoff-overlay-extra}
\end{figure}

\section{Additional Compute-Time Tradeoff Plots}
\label{app:tradeoff-plots}

Figure~\ref{fig:tradeoff-overlay-extra} overlays the plots for different thresholds. The curve moves upward as the target loss decreases, corresponding to the fact that we need more time and compute resources to achieve lower losses.

\section{A Simple Parametric Model of the Critical Batch Size}
\label{app:critical-toy}

We assume that the relationship between the number of steps $S_L(B)$ and the batch size $B$ is piecewise linear in log-log scale. This assumption is strong but may be motivated by empirical observations in existing works which insinuate this behavior \citep{zhang2019algorithmic,vyas2024soap}. 
Under this assumption, it follows that 
\begin{align*}
    \log S_L(B) = \begin{cases} - \log B + b_1 &\text{for $B \leq B_L^\star$} \\ 
    m \log B + b_2 &\text{for $B > B_L^\star$, where $-1 < m < 0$} \end{cases}
\end{align*}
with some intercepts $b_1, b_2$ such that $\log S_L(B)$ remains continuous. From $S_L(B) = T_L^\star B^{-1}$ we know $b_1 = \log T_L^\star$. We then have the following parametric forms for the number of steps and tokens:
\begin{align*}
    S_L(B) = \begin{cases} 
    e^{b_1} B^{-1} &\text{for $B \leq B_L^\star$} \\ 
    e^{b_2} B^m &\text{for $B > B_L^\star$} \end{cases} &&
    T_L(B) = \begin{cases} 
    e^{b_1} &\text{for $B \leq B_L^\star$} \\ 
    e^{b_2} B^{m+1} &\text{for $B > B_L^\star$} \end{cases} 
\end{align*}
Let the subscripts $A, M$ denote AdamW and Muon. Assuming $B_{L,A}^\star < B_{L,M}^\star$, the token ratio $R_L(B) = T_{L,A}(B) / T_{L,M}(B)$ has the following form:
\begin{align*}
    R_L(B) = \begin{cases} 
    e^{b_{1,A} - b_{1, M}}  &\text{for $B \leq B_{L,A}^\star$} \\ 
    e^{b_{2,A} - b_{1,M}} B^{m_A+1} &\text{for $B_{L,A}^\star < B \leq B_{L,M}^\star$} \\
    e^{b_{2,A} - b_{2,M}} B^{m_A-m_B} &\text{for $B > B_{L,M}^\star$} 
    \end{cases}
\end{align*}
In particular, $R_L(B)$ does not vanish in the post-critical regime iff $m_A \geq m_B$.

\vspace{-5mm}
\section{Muon Implementation}
\label{app:muon-imp}

We adapted the experimental Muon implementation on Optax.\footnote{ \url{https://github.com/google-deepmind/optax/blob/6bd761c00d839ca363bffac2584888ac797fc69d/optax/contrib/_muon.py}.}
Specifically, we modified the application of Newton-Schulz to handle the transformer weights shaped differently for different layers under \texttt{jax.lax.scan}.
We maintain the same implementation and the default hyperparameters for Newton-Schulz given by \citet{jordan2024muon}.

{\scriptsize
\begin{verbatim}
from typing import NamedTuple, Optional, Union

import chex
import jax
import jax.numpy as jnp
import optax

from optax import tree_utils as otu
from optax._src import base
from optax._src import combine
from optax._src import transform
from optax._src import utils

from collections.abc import Callable
from typing import Any, Optional, Union, Tuple
MaskOrFn = Optional[Union[Any, Callable[[base.Params], Any]]]


def orthogonalize_matrix(
      x: jax.Array,  # (batch_size, d, D)
      ns_steps: int = 5, 
      eps: float = 1e-7,
) -> jax.Array:
  a, b, c = (3.4445, -4.7750, 2.0315)
  transposed = False
  if x.shape[1] > x.shape[2]:  
    x = x.transpose(0, 2, 1)
    transposed = True

  def newton_schulz_iterator(X: jax.Array) -> jax.Array:
    A = X @ X.transpose(0, 2, 1) 
    B = b * A + c * A @ A  
    return a * X + B @ X

  x /= jnp.linalg.norm(x, axis=(1, 2), keepdims=True) + eps
  x = jax.lax.fori_loop(0, ns_steps, lambda _, x: newton_schulz_iterator(x), x)
  if transposed:
    x = x.transpose(0, 2, 1)
  return x


def orthogonalize_via_newton_schulz_scan_attn_out(
    x: jax.Array,  # (num_heads, num_layers, dim_head, dim)
    ns_steps: int = 5,
    eps: float = 1e-7,
) -> jax.Array:
  (num_heads, num_layers, dim_head, dim) = x.shape
  x = x.transpose(1, 0, 2, 3)  # (num_layers, num_heads, dim_head, dim)
  x = x.reshape(num_layers, num_heads * dim_head, dim)
  x = orthogonalize_matrix(x, ns_steps, eps)
  x = x.reshape(num_layers, num_heads, dim_head, dim)
  x = x.transpose(1, 0, 2, 3)  # (num_heads, num_layers, dim_head, dim)
  return x


def orthogonalize_via_newton_schulz_scan_attn_qkv(
    x: jax.Array,  # (dim, num_layers, num_heads, dim_head)
    ns_steps: int = 5,
    eps: float = 1e-7,
) -> jax.Array:
  (dim, num_layers, num_heads, dim_head) = x.shape
  x = x.transpose(1, 0, 2, 3)  # (num_layers, dim, num_heads, dim_head)
  x = x.reshape(num_layers, dim, num_heads * dim_head) 
  x = orthogonalize_matrix(x, ns_steps, eps)
  x = x.reshape(num_layers, dim, num_heads, dim_head) 
  x = x.transpose(1, 0, 2, 3)  # (dim, num_layers, num_heads, dim_head)
  return x


def orthogonalize_via_newton_schulz_scan_mlp(
    x: jax.Array,  # (dim, num_layers, dim_hidden)
    ns_steps: int = 5,
    eps: float = 1e-7,
) -> jax.Array:
  x = x.transpose(1, 0, 2)  # (num_layers, dim, dim_hidden)
  x = orthogonalize_matrix(x, ns_steps, eps)
  x = x.transpose(1, 0, 2)  # (dim, num_layers, dim_hidden)
  return x

    
def apply_orthogonalization_logic(x, path, ns_steps, eps, base_scale: float):
    path_str = "/".join(path).lower()

    # Name-based logic, may need to change for more robust handling.
    if "mlp" in path_str:
        assert x.ndim == 3
        (d1, _, d2) = x.shape
        x_orth = orthogonalize_via_newton_schulz_scan_mlp(x, ns_steps, eps)
    elif "attention" in path_str and "out" not in path_str:
        assert x.ndim == 4
        (d1, _, num_heads, dim_head) = x.shape
        d2 = num_heads * dim_head
        x_orth = orthogonalize_via_newton_schulz_scan_attn_qkv(x, ns_steps, eps)
    elif "attention" in path_str and "out" in path_str:
        assert x.ndim == 4
        (num_heads, _, dim_head, d1) = x.shape
        d2 = num_heads * dim_head
        x_orth = orthogonalize_via_newton_schulz_scan_attn_out(x, ns_steps, eps)
    else:
        raise ValueError(f"Unidentifiable path for Muon: {path_str}")

    scale = base_scale * jnp.sqrt(jnp.maximum(d1, d2)) if base_scale > 0.0 else 1.0  
    return x_orth * scale


def orthogonalize_tree(
    pytree: Any,
    ns_steps,
    eps,
    base_scale, 
    path: Tuple[str, ...] = (),
) -> Any:
    if isinstance(pytree, optax.MaskedNode):
        return pytree
    elif isinstance(pytree, dict):
        return {k: orthogonalize_tree(v, ns_steps, eps, base_scale, path + (k,)) for k, v in pytree.items()}
    else:
        return None if pytree is None else apply_orthogonalization_logic(pytree, path, ns_steps, eps, base_scale)


class MuonState(NamedTuple):
  mu: base.Updates


def scale_by_muon(
    momentum: float = 0.95,
    *,
    mu_dtype: Optional[chex.ArrayDType] = None,
    nesterov: bool = True,
    eps: float = 1e-7,
    ns_steps: int = 5,
    base_scale: float = 0.2,
) -> base.GradientTransformation:
  mu_dtype = utils.canonicalize_dtype(mu_dtype)

  def init_fn(params):
    mu = otu.tree_zeros_like(params, dtype=mu_dtype)  # First moment
    return MuonState(mu=mu)

  def update_fn(updates, state, params=None):
    del params
    
    def momentum_grad(grad, buf):
      return grad + momentum * buf if grad is not None else None

    new_mu = jax.tree.map(momentum_grad, updates, state.mu, is_leaf=lambda x: x is None)
    mu = jax.tree.map(momentum_grad, updates, new_mu, is_leaf=lambda x: x is None) if nesterov else new_mu
    mu_orth = orthogonalize_tree(mu, ns_steps, eps, base_scale)

    return mu_orth, MuonState(mu=otu.tree_cast(new_mu, mu_dtype))
  
  return base.GradientTransformation(init_fn, update_fn)


def my_param_labels(params):
    def recurse(subtree, path=()):
        if isinstance(subtree, dict):
            return {k: recurse(v, path + (k,))  for k, v in subtree.items()}
        else:
            if any(('norm' in k or 'logits' in k or 'embedding' in k) for k in path):
                return 'adam'
            else:
                return 'muon'
    return recurse(params)


def muon(
    learning_rate: base.ScalarOrSchedule,
    momentum: float = 0.95,
    *,
    mu_dtype: Optional[chex.ArrayDType] = None,    
    nesterov: bool = True,
    eps: float = 1e-7,
    ns_steps: int = 5,
    base_scale: float = 0.2,
    adam_b1: float = 0.95,
    adam_b2: float = 0.95,
    adam_eps: float = 1e-8,    
    weight_decay: float = 0.1,
    mask: MaskOrFn = None,
) -> base.GradientTransformation:
  return combine.chain(
    combine.multi_transform(
      transforms={
        'muon': scale_by_muon(
          momentum=momentum,
          mu_dtype=mu_dtype,
          nesterov=nesterov,
          eps=eps,
          ns_steps=ns_steps,
          base_scale=base_scale,
        ),
        'adam': transform.scale_by_adam(
          b1=adam_b1,
          b2=adam_b2,
          eps=adam_eps,
          eps_root=0,
          mu_dtype=mu_dtype,
          nesterov=nesterov,
        )
      },
      param_labels=lambda params: my_param_labels(params),
    ),
    transform.add_decayed_weights(weight_decay, mask), 
    transform.scale_by_learning_rate(learning_rate),       
  )
\end{verbatim}
}

\section{Parameterization of Telescoping and Distribution of Losses}
\begin{figure}[H]
    \centering
    \includegraphics[width=1\linewidth]{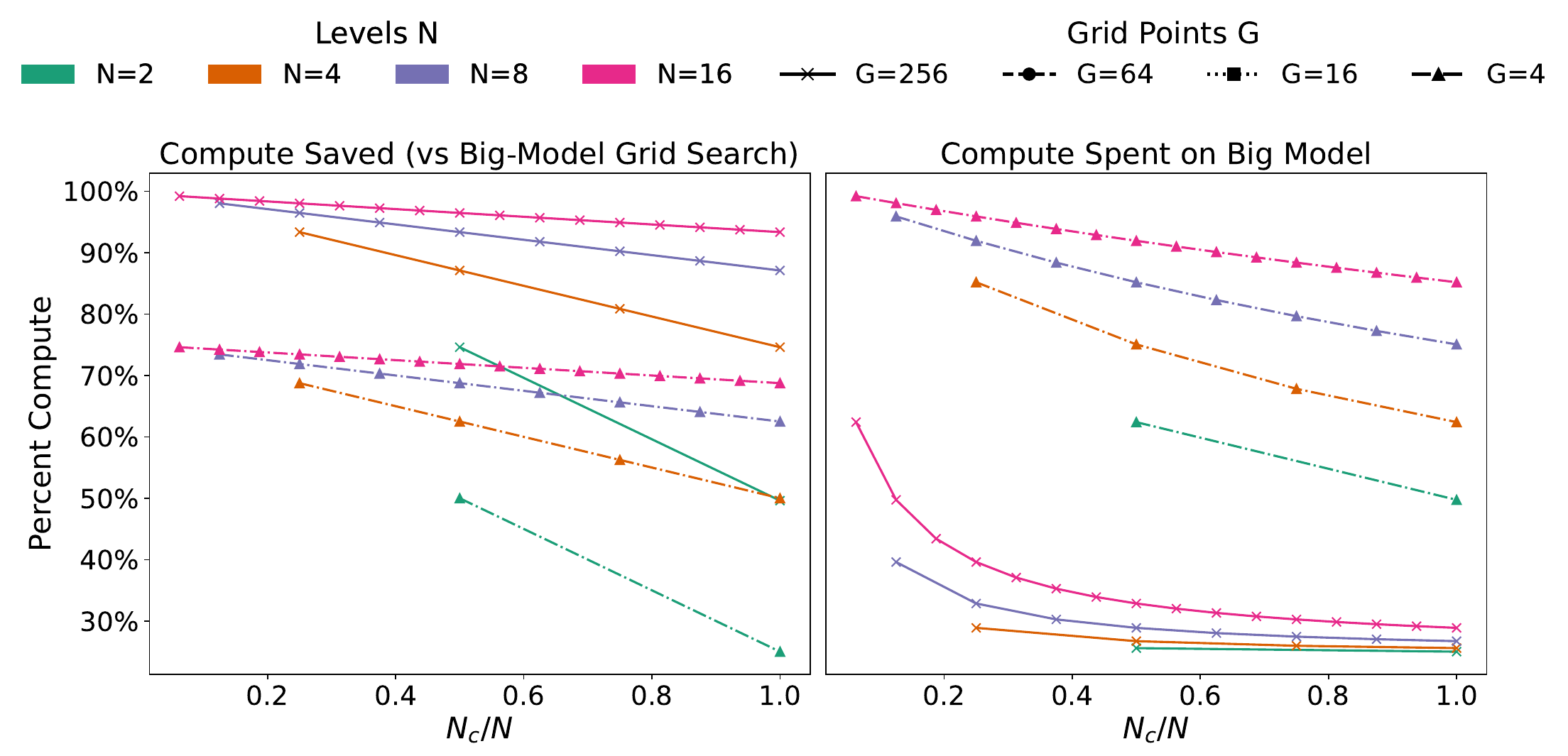}
    \caption{$N_c$ is the calibration level of the telescope - the level corresponding to the largest width we perform the telescoping up to. Two figures of merit: (left) the percentage of compute saved using telescoping muP as opposed to performing a full grid search on the full model size, for $G$ grid points, and $N$ levels the base model width. Savings are largest for large grid sweeps ($G$  large), large models ($N$ large), and small truncation ratios ($N_c/N$ small). (right) The percentage of the total compute spent on the final model run. Because we perform the same amount of compute at each stage of the telescope, this is maximized for small $N_c/N$, following the procedure suggested by \cite{yang2022tensor}. This measures a form of exploration-exploitation trade off.}
    \label{fig:telescope-analysis}
\end{figure}
\begin{figure}[H]
    \centering
    \includegraphics[width=.8\linewidth]{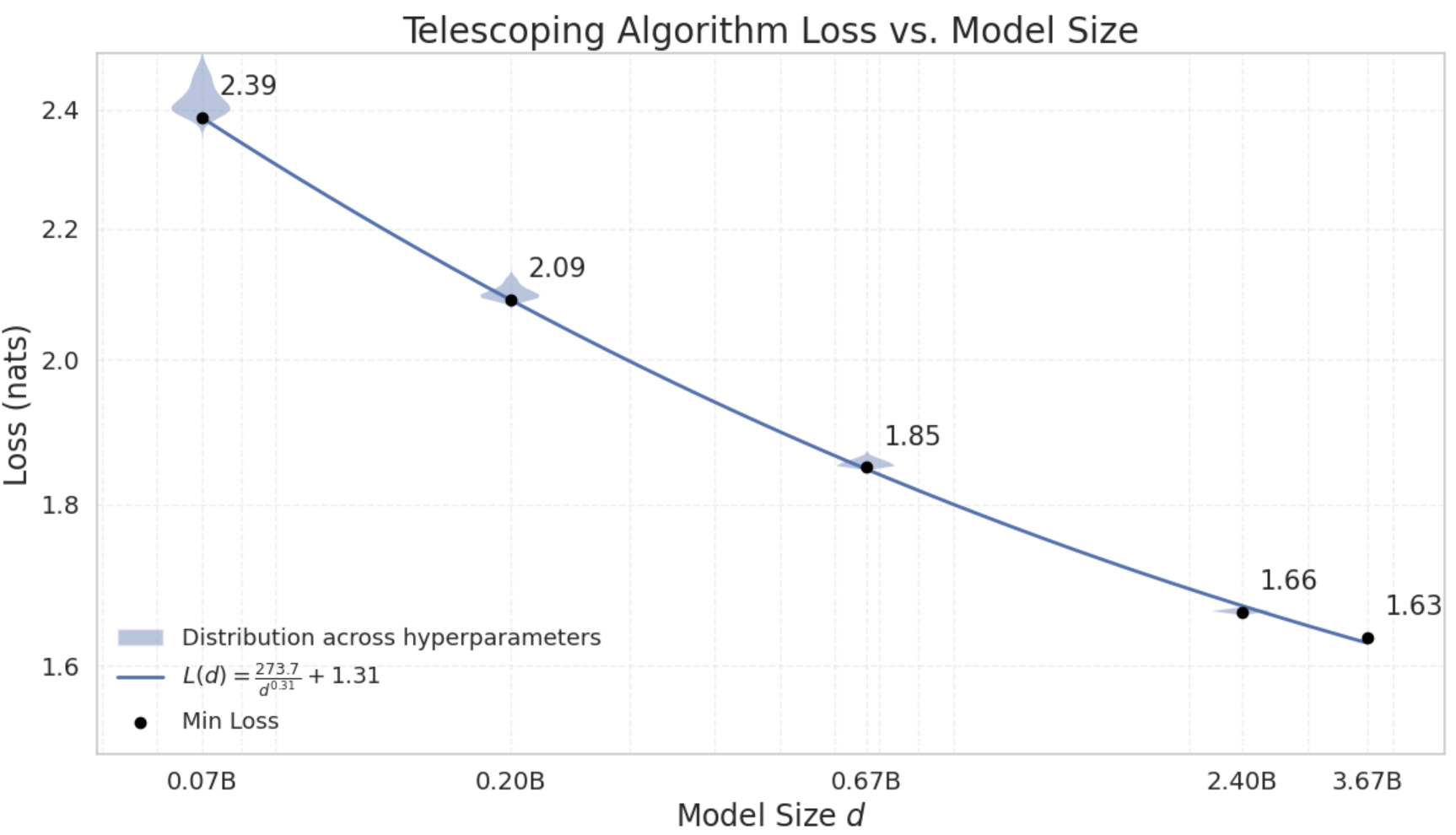}
    \caption{The distribution of loss values at each level of the telescope in Fig.~\ref{fig:telescope},  truncated at 2.5 to omit outliers from the initial broad sweep. We note that as the telescope level increases, the distribution becomes tighter around its mean, demonstrating that we are producing a precise estimate of the optimal hyperparameters, and the loss continues to decrease demonstrating that we maintain accuracy. The minimum loss values at each point fit well ($R^2\approx 1$) to a Chinchilla style loss function ($L(d) = A/d^{\alpha} + E$) with parameter count ($d$) exponent of $\alpha=0.31$, close to the $0.34$ in \cite{hoffmann2022training}. Because the tokens are held constant across all scales, the second loss term ($B/T^{\beta}$, for $T$ tokens and $B$ and $\beta$ fit parameters in \cite{hoffmann2022training}) is the same for all models and hence included in the fit constant ($E= 1.31$) here.}
    \label{fig:loss_values}
\end{figure}
\section{The Maximal Update Parameterization}\label{app:mup}

The maximal update parameterization (muP) enables scale-invariant neural network training by prescribing width-dependent initialization and learning rate scalings. A modern interpretation of muP is presented in terms of spectral norms and linear algebra \cite{yang2023spectral}. Let $W_\ell \in \mathbb{R}^{n_\ell \times n_{\ell-1}}$ denote the weight matrix for layer $\ell$. We require that a vector $h_{\ell-1}$ of norm $\Theta(\sqrt{n_{\ell-1}})$ maps to $h_\ell := W_\ell h_{\ell-1}$ with norm $\Theta(\sqrt{n_\ell})$. This is ensured if the spectral norm satisfies:
\[
\|W_\ell\|_* = \Theta\left(\sqrt{\frac{n_\ell}{n_{\ell-1}}}\right),
\]
which can be understood as maintaining vector norms throughout the network. The spectral norm controls the largest singular value of each layer, hence for an input with typical norm $\sqrt{n_0}$, each layer will in turn have typical norm $\sqrt{n_{\ell}}$, as expected from a vector of size $n_{\ell}$ whose entries are $\Theta(1)$. This condition, interpretable as \(\sqrt{\mathrm{fan\_out}/\mathrm{fan\_in}}\), enforces stable propagation of activations and gradients. Under gradient descent, updates $\Delta W_\ell$ with the same scaling ensure that $\Delta h_\ell = \Delta W_\ell h_{\ell-1} + \dots$ remains order-one.

Contrast this approach with standard arguments (e.g., \cite{yang2019scaling}) which enforce this kind of scale invariance by ensuring stable forward and backward dynamics by requiring:
\begin{equation}\label{eq:mup-constraints}
\mathbb{E}\bigl[W_{\ell}^{i_1j_1}W_{\ell}^{i_2j_2}\bigr] \sim \frac{1}{n^p},
\qquad 
\eta \sim \frac{1}{n^q},
\end{equation}
where $n$ is the layer width, $i_1, i_2, j_1, j_2$ are indices and $p, q$ are chosen to avoid exploding or vanishing signals. That this can be accomplished is non-obvious, and is the subject of a substantial body of literature \cite{yang2019scaling, yaida2022meta, halverson2024physics}. While generally less interpretable, this description of muP is convenient for two reasons. First, it generally makes clear the effect of re-parameterizations. Because the learning dynamics have three free parameters (initialization variance, learning rate and initialization scale) but only two constraints (Eq.~\ref{eq:mup-constraints}), one degree of freedom remains to choose convenient parameterizations (e.g., to allow weight tying between the embedding and unembedding). The spectral and dynamical conditions we have mentioned can also be shown to agree: the spectral condition guarantees forward and backward stability and matches earlier $mu$P tables \cite{yang2019scaling, yang2022tensor}, and specifically Table~\ref{table:params}. Second, this analysis allows a simple accounting for the possible sources of error in using muP, which we will look at in the following section.

\section{Hyperparameter Drift under MuP Scaling}\label{sec:taylor}

While muP yields approximate scale-invariance, finite-width effects introduce predictable hyperparameter drift. Let $f(x, n)$ be the output of a width-$n$ network under hyperparameter $x$ (e.g., learning rate), and $f_0(x)$ its infinite-width limit. Then:
\[
f(x, n) = f_0(x) + \frac{f_1(x)}{n} + O\!\left(\frac{1}{n^2}\right).
\]
Minimizing a loss $\mathcal{L}(f(x,n))$ leads to an optimum $x^\star(n)$ that drifts from its infinite-width counterpart $x^\star$:
\[
x^\star(n) = x^\star - \frac{\alpha}{n} + O\!\left(\frac{1}{n^2}\right),
\]
where $\alpha$ depends on $f_0$, $f_1$, and $\mathcal{L}$. Expanding the loss:
\[
\ell(x,n) = \mathcal{L}(f(x,n)) 
= \ell_0(x) + \frac{1}{n} \ell_1(x) + O\!\left(\frac{1}{n^2}\right),
\quad \ell_1(x) = \mathcal{L}'(f_0(x))f_1(x).
\]
Imposing optimality yields:
\[
\alpha = -\frac{\ell_1'(x^\star)}{\ell_0''(x^\star)} 
= -\frac{
\mathcal{L}''(f_0)\,f_0'\,f_1 + \mathcal{L}'(f_0)\,f_1'
}{
\mathcal{L}'(f_0)\,f_0'' + \mathcal{L}''(f_0)\,[f_0']^2
}.
\]
At an extremum where one term in the denominator vanishes, we recover simplified drift laws:
\[
\alpha =
\begin{cases}
\displaystyle \frac{f_1(x^\star)}{f_0'(x^\star)}, & \text{if } \mathcal{L}'(f_0(x^\star)) = 0,\\[1em]
\displaystyle \frac{f_1'(x^\star)}{f_0''(x^\star)}, & \text{if } f_0'(x^\star) = 0.
\end{cases}
\]
Thus, hyperparameter optima shift by $O(1/n)$ due to finite-width corrections, with the scale and direction of the drift governed by the structure of $f_0$ and $f_1$. This formalizes and quantifies the empirical observation that minima shift during width scaling (e.g., see Appendix of \cite{everett2024scaling}), and provides the key error source in muP-based hyperparameter transfer.

\end{document}